%% file: main.tex
\documentclass[10pt,twocolumn,letterpaper]{article}

\usepackage{iccv}              
\usepackage [accsupp]{axessibility} 
\usepackage{graphicx} 
\usepackage{rotating} 
\usepackage{tabularx} 
\usepackage{multirow}
\usepackage{makecell}
\usepackage{tikz} 
\usepackage{pgfplots} 
\usepackage{adjustbox} %
\usepackage{amssymb} 
\usepackage{xcolor}   
\usepackage{amsmath}

\usepackage{pifont}
\definecolor{mygray}{gray}{.95}
\definecolor{mycyan}{cmyk}{.1,0,0,0}
\definecolor{darkgreen}{rgb}{0.0, 0.5, 0.0}

\usepackage[dvipsnames]{xcolor}

\newcommand{\cmarkg}{\ding{51}}%
\newcommand{\xmarkg}{\textcolor{lightgray}{\ding{55}}}%


\definecolor{iccvblue}{rgb}{0.21,0.49,0.74}
\usepackage[pagebackref,breaklinks,colorlinks,allcolors=iccvblue]{hyperref}

\def\paperID{7925} 
\def\confName{ICCV}
\def\confYear{2025}

\title{\textit{RoboTron-Mani}: All-in-One Multimodal Large Model for Robotic Manipulation}

\author{Feng Yan\thanks{Equal contribution. Email: bphengyan@163.com} \quad Fanfan Liu\footnotemark[1] \quad Yiyang Huang \quad Zechao Guan \quad Liming Zheng \\
\quad Yufeng Zhong \quad Chengjian Feng \quad Lin Ma\thanks{Corresponding author.}
\\ \\
{ Meituan}\\
\url{https://github.com/EmbodiedAI-RoboTron/RoboTron-Mani}
}

\pgfplotsset{compat=1.18}

\newcommand{\iRoboMM}{\textit{RoboTron-Mani}}  

\newcommand{\tRoboMM}{\text{RoboTron-Mani}}

\begin{document}
\maketitle

\begin{abstract}

{Recently}, robotics has advanced significantly through the integration of larger models and large-scale datasets. However, challenges remain in applying these models to 3D spatial interactions and managing data collection costs. To address these issues, we propose 
the multimodal robotic manipulation model \textbf{\iRoboMM{}} and the comprehensive dataset \textbf{RoboData}. \iRoboMM{}, on one hand,  enhances 3D perception through camera parameters and occupancy supervision.  On the other hand, it further incorporates Modality-Isolation-Mask and multimodal decoder blocks based on OpenFlamingo, improving modality fusion and fine-grained perception. \textit{RoboData} integrats several publicly-available datasets, achieving the first fusion of multi-view images, camera parameters, depth maps, actions, and space alignment, which facilitates comprehensive learning from diverse robotic datasets and offers one complete evaluation system. Trained on \textit{RoboData}, \iRoboMM{} is the first generalist policy that surpasses expert models, enabling simultaneous evaluation of all tasks across multiple datasets, rather than being limited to specific data or task selections. Specifically, \iRoboMM{} boosts manipulation performance by increasing the average sequence length on CALVIN from 1.7 to 3.5, enabling cross-embodiment generalization, and achieving state-of-the-art results on both simulated and real-world datasets.

\end{abstract}

\begin{figure}[!t]
    \centering 
    \includegraphics[width=1.0\linewidth]{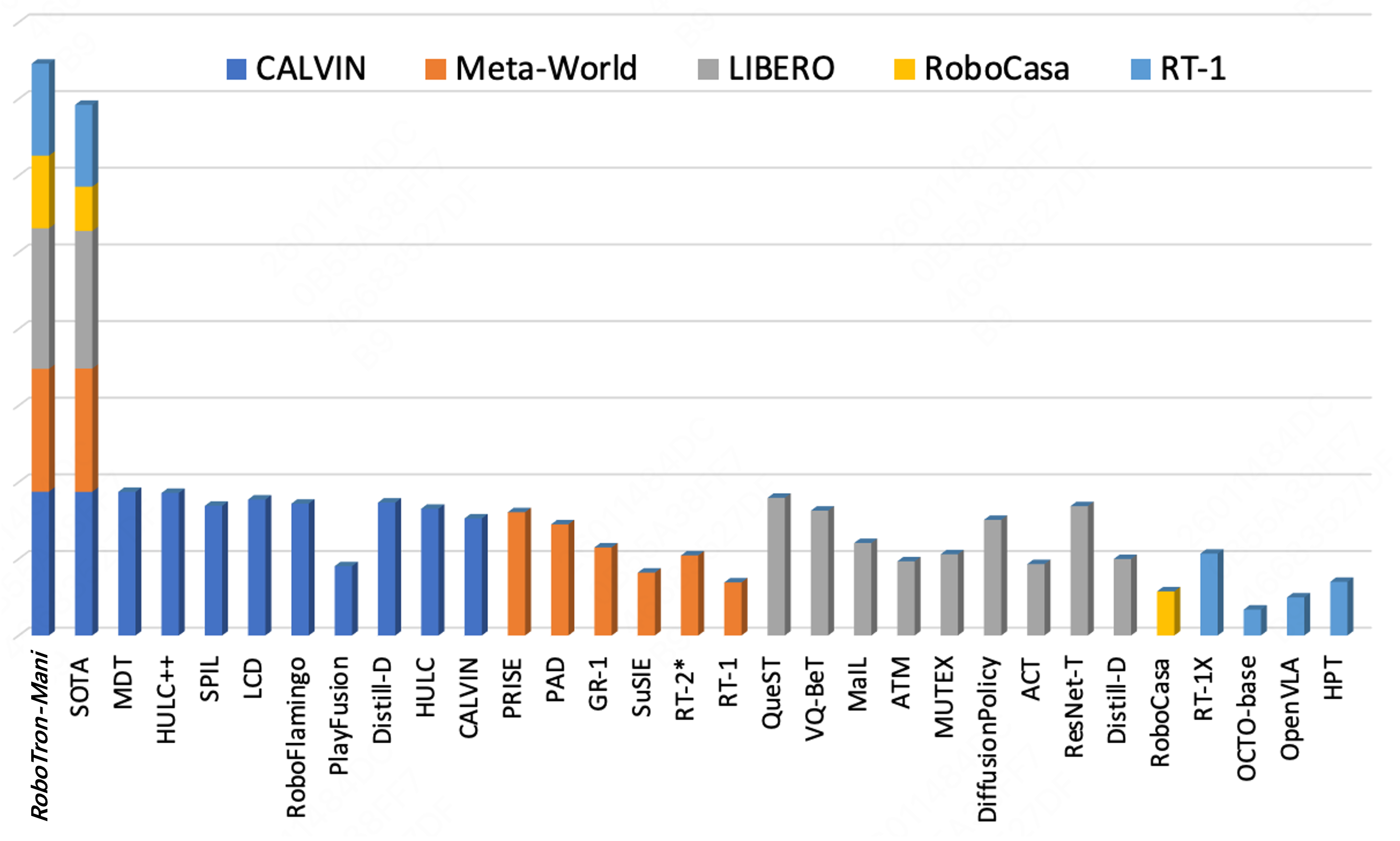} 
    \vspace{-6mm}
    \caption{Stacked bar chart depicting the performance of various models across five datasets. The ``SOTA" label represents the best results achieved by specialized models for each dataset. Notably, \iRoboMM{} is the only generalist policy evaluated on multiple datasets, demonstrating competitive performance relative to ``SOTA". For more details, please refer to Section~\ref{sec:sota}.}
    \label{fig:results}
    \vspace{-5mm}
\end{figure}

\section{Introduction}
In recent years, machine learning has experienced profound advancements, from the advent of CLIP~\cite{radford2021learning, zhao2023learning} to the progression of foundational models like the GPT series~\cite{brown2020language, achiam2023gpt, bubeck2023sparks}, Llama~\cite{touvron2023llama, touvron2023llama2}, LLaVA~\cite{liu2024visual}, and Flamingo~\cite{alayrac2022flamingo, awadalla2023openflamingo}. These strides are largely due to larger transformer-based architectures and the utilization of ``internet-scale'' datasets~\cite{liu2024llava, chen2023sharegpt4v, chen2024allava}. These innovations have not only extended the frontiers of natural language processing~\cite{chowdhary2020natural} and computer vision~\cite{guo2022attention, ren2024groundedsamassemblingopenworld} but also galvanized researchers to integrate these models into Embodied Artificial Intelligence (EAI)~\cite{ollero2021past}, thereby enabling more complex and varied tasks in real-world environments.

\begin{figure*}[!ht]
    \centering 
    \includegraphics[width=1.0\linewidth]{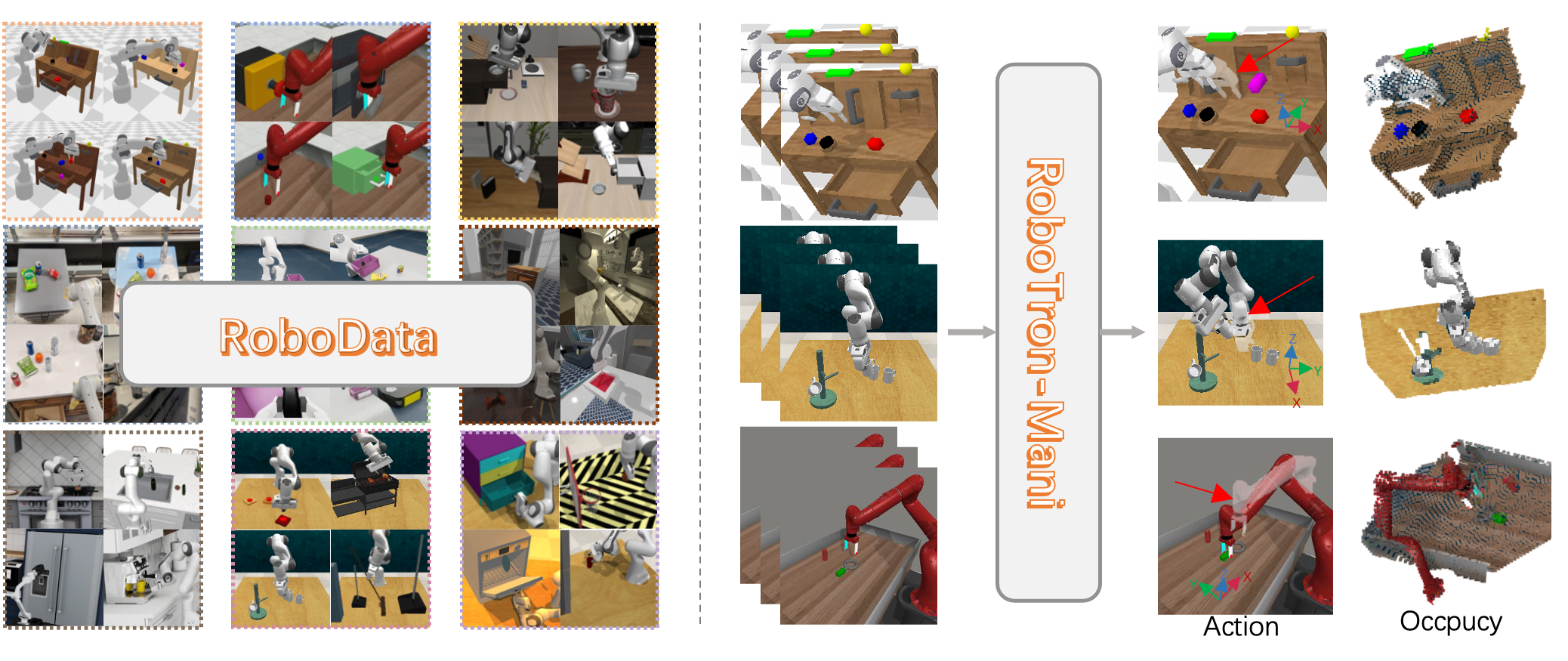} 
    \vspace{-8mm}
    \caption{\textbf{Left:} \textit{RoboData} integrates multiple diverse and complex datasets (CALVIN~\cite{mees2022calvin}, Meta-World~\cite{yu2020meta}, LIBERO~\cite{liu2024libero}, RT-1~\cite{brohan2022rt}, RoboCAS~\cite{zheng2024robocas}, ManiSkill2~\cite{gu2023maniskill2}, RoboCasa~\cite{nasiriany2024robocasa}, RLBench~\cite{james2020rlbench}, and Colosseum~\cite{pumacay2024colosseum}), covering various robot embodiments, environments, and task types, resulting in a unified dataset with standardized input and output spaces. \textbf{Right:} \iRoboMM{} features comprehensive 3D perception capabilities, flexible multimodal outputs, and significantly enhances the robotic manipulation generalization capabilities.}
    \label{fig:robomm}
    \vspace{-6.mm}
\end{figure*}

In terms of modeling, there has been a gradual shift from single-task or single-dataset learning~\cite{pmlr-v205-shridhar23a, mees2023grounding, zhao2023learning} towards transfer learning approaches~\cite{brohan2023rt, kim2024openvla, liu2024robouniview, wang2024scaling, team2024octo, huang2024corkienablingrealtimeembodied}. These models leverage robust foundational models that are pre-trained on extensive ``internet-scale'' datasets or diverse data sources. Subsequently, they undergo fine-tuning on specific robotic datasets to produce precise control actions.
On the data front, researchers collect data through various means to augment models. For example, the Open X-Embodiment~\cite{padalkar2023open} amalgamates diverse datasets containing vision-language-action pairs, whereas the RH20T~\cite{fang2023rh20t} gathers data via teleoperation. 
Despite the impressive robustness of these efforts, they still encounter significant challenges in practical applications.
First, ~\textbf{is the direct application of multimodal models to EAI the optimal solution?} Robots should interact with the physical 3D space. However, current multimodal models predominantly focus on 2D image understanding and generation, 
limiting their practical applicability, especially in physical-world interaction. 
Second, ~\textbf{is it essential to address the cost and efficiency of dataset construction?} For instance, collecting around 130,000 episodes from the RT-1~\cite{brohan2022rt} took 17 months. Therefore, it is imperative to integrate as many existing multi-platform, multi-robot datasets from the industry as possible to address this urgent issue.

For tackling the aforementioned challenges, this paper introduces \iRoboMM, the multimodal large model for robotic manipulation, and \textit{RoboData}, the comprehensive dataset integrating datasets across various platforms and robots for evaluation and training purposes. Figure~\ref{fig:robomm} shows the framework and capabilities of our dataset platform and multimodal robot model.

In terms of modeling, \iRoboMM{} combines camera parameters and occupancy supervision to enhance 3D environmental perception. Additionally, leveraging large language models like OpenFlamingo~\cite{awadalla2023openflamingo}, we design the plug-and-play efficient Modality-Isolation-Mask (MIM), which flexibly introduces multimodal supervision. This gives the model fine-grained perception and leverages large-scale internet data.

On the data side, while Open X-Embodiment~\cite{padalkar2023open} integrates multiple datasets, it lacks critical information such as multi-view images, camera parameters, and depth maps, making it more suitable for 2D multimodal training. Additionally, the lack of data space alignment prevents the robot's 6D pose from being consistent across different datasets, yielding the results in Open X-Embodiment~\cite{padalkar2023open} Table 1, where direct data fusion results in RT-1-X underperforming RT-1. In contrast, \textit{RoboData} addresses these limitations by integrating a wide range of well-known industry datasets, including simulated datasets (CALVIN~\cite{mees2022calvin}, Meta-World~\cite{yu2020meta}, LIBERO~\cite{liu2024libero}, Robomimic~\cite{robomimic2021}, RoboCAS~\cite{zheng2024robocas}, ManiSkill2~\cite{gu2023maniskill2}, RoboCasa~\cite{nasiriany2024robocasa}, RLBench~\cite{james2020rlbench}, and Colosseum~\cite{pumacay2024colosseum}), and RT-1~\cite{brohan2022rt}. We dedicate hundrads of person-days to collect and organize these data, supplementing missing modalities such as depth maps and camera parameters. Moreover, \textit{RoboData} unifies the input and output spaces across different robots and platforms, ensuring consistent perception and action representation, and enabling integrated learning from diverse simulated and real-world datasets.

\begin{figure*}[!ht]
    \centering
    \includegraphics[width=1.0\linewidth]{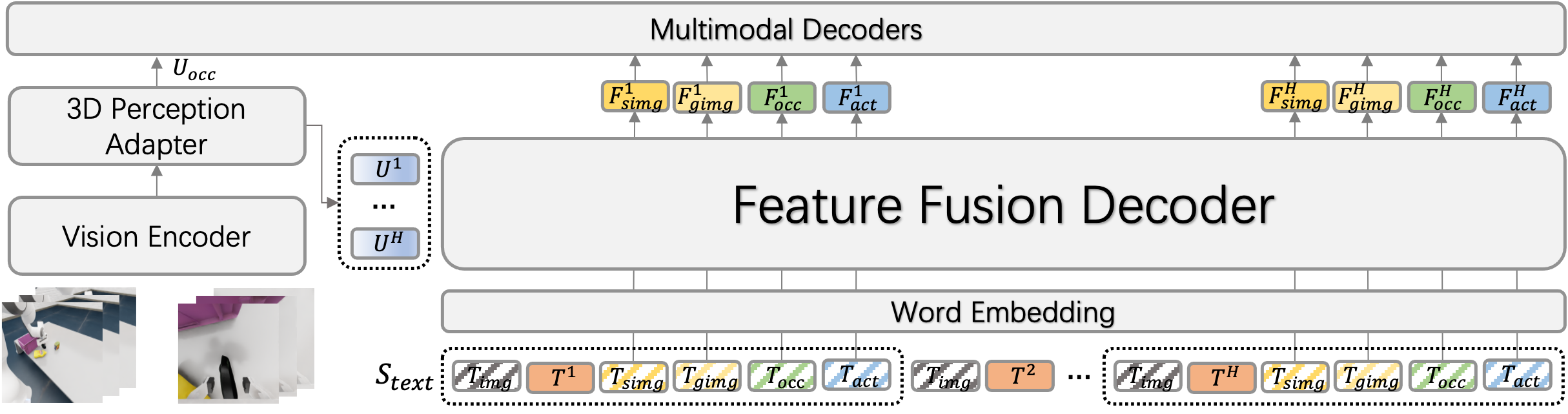} 
    \caption{Architecture of \iRoboMM. Vision Encoder extracts multi-view features. 3D Perception Adapter leverages occupancy supervision to unify features and enhance spatial perception. Feature Fusion Decoder based on LLMs merges text and visual information. Multimodal Decoders enhance fine-grained perception and understanding through multimodal outputs.}
    \label{fig:main}
    \vspace{-4.5mm}
\end{figure*}

Extensive experiments demonstrate that the various components of \iRoboMM{} significantly improve performance in robotic manipulation tasks, enhancing the average sequence length on the CALVIN~\cite{mees2022calvin} benchmark from 1.7 to 3.5. Additionally, \iRoboMM{} ensures cross-embodiment capabilities and the first generalist policy that surpasses expert models, enabling simultaneous evaluation of all tasks across multiple datasets, rather than being limited to specific data or task selections, as shown in Figure~\ref{fig:results}. This paper underscores the critical role of advanced models and curated datasets in advancing robotics, highlighting their potential to drive significant improvements in robotic performance. In summary, \textbf{\textit{RoboData} aims to provide the research community with a comprehensive and fair evaluation system. And \iRoboMM{} is the first generalist policy to incorporate training and testing across multiple datasets.}

\section{\iRoboMM}

\subsection{Preliminary}

Multimodal Large Language Models (MLLMs) typically consist of three main components: the modality encoder (Enc), the adapter (Adapter), and the large language model (LLM), mathematically expressed as follows:
\begin{align}
    O_T &= \text{MLLM}(I, T) \notag \\
        &= \text{LLM}\left(\text{Adapter} \left(\text{Enc}(I)\right), \text{WE}(T)\right).
\end{align}
Here, \text{WE} denotes the word embedding layer.
The modality encoder transforms inputs from single modality into appropriate representations. For instance, the image encoder extracts features \(F_I\) from input images \(I\). Common visual encoders like CLIP~\cite{radford2021learning} are pre-trained on image-text pairs, aligning visual and textual semantics for easier integration with LLMs.
The adapter maps features from visual and other modalities into inputs \(U\) that the LLM can understand. For example, Blip2~\cite{li2023blip} uses Q-Former for feature interaction; LLaVA~\cite{liu2024visual} employs MLPs to align visual features with text features.
The large language model is the core component of our framework, referred to in this paper as the Feature Fusion Decoder. It typically employs auto-regressive models such as LLaMA~\cite{touvron2023llama} or GPT~\cite{achiam2023gpt}, as well as cross-attention models like Flamingo~\cite{alayrac2022flamingo} or LLaMA3.2~\cite{meta2024llama}. This model fuses the feature representations \(U\) with text features \(F_T\) extracted from the word embedding layer to generate the final textual output \(O_T\). This integration of features enhances the model's ability to produce contextually relevant responses.

\begin{figure*}[ht]
    \centering
    \begin{subfigure}{0.21\textwidth}
        \centering
        \includegraphics[width=\linewidth]{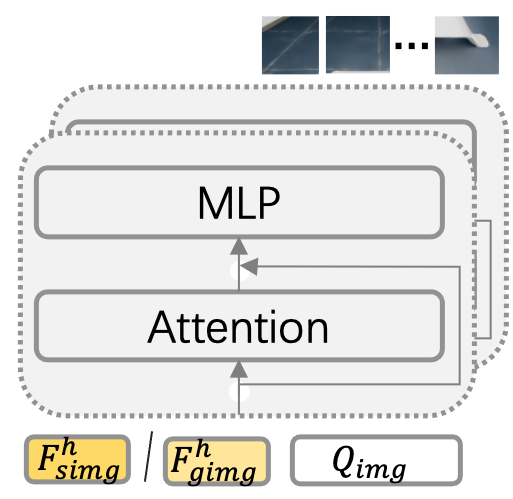}  
        \caption{Image Decoder}
        \label{fig:sub_imgd}
    \end{subfigure}
    \hfill
    \begin{subfigure}{0.4\textwidth}
        \centering
        \includegraphics[width=\linewidth]{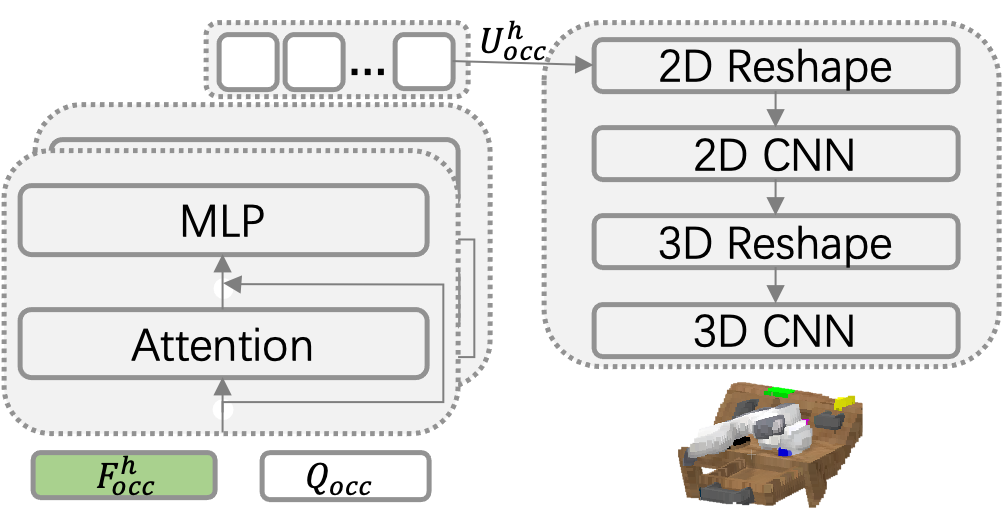}  
        \caption{Occupancy Decoder}
        \label{fig:sub_occd}
    \end{subfigure}
    \hfill
    \begin{subfigure}{0.33\textwidth}
        \centering
        \includegraphics[width=\linewidth]{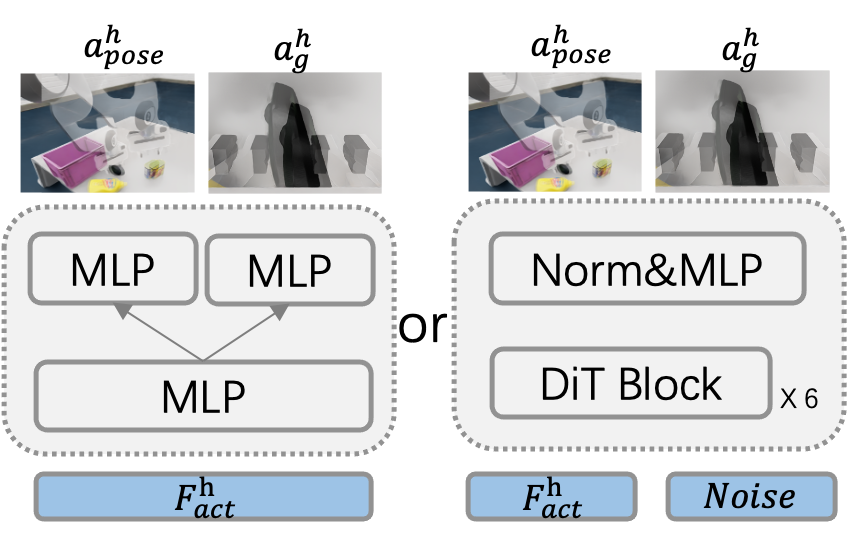}  
        \caption{Action Decoder (MLP or DiT)}
        \label{fig:sub_actd}
    \end{subfigure}
    \vspace{-2mm}
    \caption{Overview of the multimodal decoders. (a) Image Decoder, (b) Occupancy Decoder, and (c) Action Decoder. Each decoder processes input features through a series of Multi-Layer Perceptrons (MLPs), attention mechanisms, and convolutional neural networks (CNNs) to generate appropriate output representations.}
    \label{fig:mainfig}
    \vspace{-5.5mm}
\end{figure*}


\subsection{Architecture}

For robotic manipulation tasks based on language instructions \(T\), which typically rely on historical frames \(I\) from \(N\) perspectives at \(H\) time steps, the task can be mathematically expressed as \(O_A = \Theta(I, T)\), where \(I \in \mathbb{R}^{H \times N \times H \times W \times 3}\). Integrating the principles of MLLMs, this paper proposes the novel multimodal robotic manipulation model named \iRoboMM, as show Figure~\ref{fig:main}, which has the capability of 3D environment perception and handles multimodal inputs (text \(T\), vision \(I\), camera parameters \(Cam\)) and outputs (actions \(O_A\), images \(O_I\), occupancy \(O_o\)):
\begin{align}
    (O_A, [O_I, O_O]) = \tRoboMM(T, I, Cam).
\end{align}
\iRoboMM{} consists of the following key components: (1) Vision Encoder  extracts observation features \(F_I^{h,n}\) from \(H\) time steps and \(N\) perspectives. (2) 
The 3D Perception Adapter enhances spatial understanding by incorporating camera parameters.
(3) Feature Fusion Decoder Based on Large Language Models merges text and visual information to output multimodal features, with the use of Modality-Isolation-Mask~(MIM) increasing the flexibility of modality fusion. (4) Multimodal Decoder enhances the model's fine-grained perception and understanding through multimodal outputs. Notably, thanks to MIM, \(O_I, O_o\) are optional outputs.

\textbf{3D Perception Adapter.} We employ UVFormer from RoboUniView~\cite{liu2024robouniview}, which is a simple yet powerful 3D environment perception model. UVFormer takes image features \(X^h = \{F_I^{h,n}\}_n^N\), camera parameters \(Cam^h = \{Cam^{h,n}\}_n^N\), and learnable unified view queries \(Q\) as inputs and outputs the unified view representation \(U_I^h\):
\begin{align}
    U_I^h = \text{UVFormer}(Q, X^h, Cam^h).
\end{align}
Here, \(Q = \{Pos, Emb\}\), \(Pos \in \mathbb{R}^{L \times B \times 3P}\) and \(Emb \in \mathbb{R}^{L \times B \times C}\) represent the positions and learnable features of the queries, respectively. \(L\), \(B\), and \(P\) define the spatial shape of the 3D grid within the operational space of the robot. Specifically, \(Emb_{l,b} \in \mathbb{R}^C\) is responsible for the corresponding pillar cell area in unified view space. \(U_I^h \in \mathbb{R}^{L \times B \times C}\) is the unified view representation, containing all relevant information in the \(L \times B \times P\) 3D grid.

\textbf{Feature Fusion Decoder.} Due to the need to support multi-frame or video inputs, we abandon the Auto-Regressive (AR) mechanism used in LLaVA~\cite{liu2024llava} and adopt OpenFlamingo~\cite{awadalla2023openflamingo} with cross-attention as the Feature Fusion Decoder. It integrates unified visual representations with language and other modality placeholders through cross-attention layers.

(a) To support multimodal output, we first construct the text sequence \(T'\), which includes text and multiple modality read-out tokens:
\begin{align}
T' = \{ &[T_{\text{img}}, T^h, T_{\text{simg}}, T_{\text{gimg}}, T_{\text{occ}}, T_{\text{act}}]\}_h^H.
\end{align}
Here, \(T'\in \mathbb{R}^{\sum_h^H (1+L^h+8*3+1)}\), \(T_{\text{simg}}, T_{\text{gimg}}, T_{\text{occ}}, T_{\text{act}}\) represent read-out tokens for static images, wrist images, occupancy, and actions, respectively. \(L^h\) represents the length of \(T^h\). \(T_{\text{img}}\) is used to indicate the position of the original image. \(T_{\text{simg}}, T_{\text{gimg}}, T_{\text{occ}}\) each use 8 tokens.
We then feed the constructed text sequence into the word embedding layer to obtain text features:
\begin{align}
F_{T} = \text{WE}(T').
\end{align}

\begin{figure}
    \centering
    \begin{minipage}{0.22\textwidth}
        \centering
        \includegraphics[width=\textwidth]{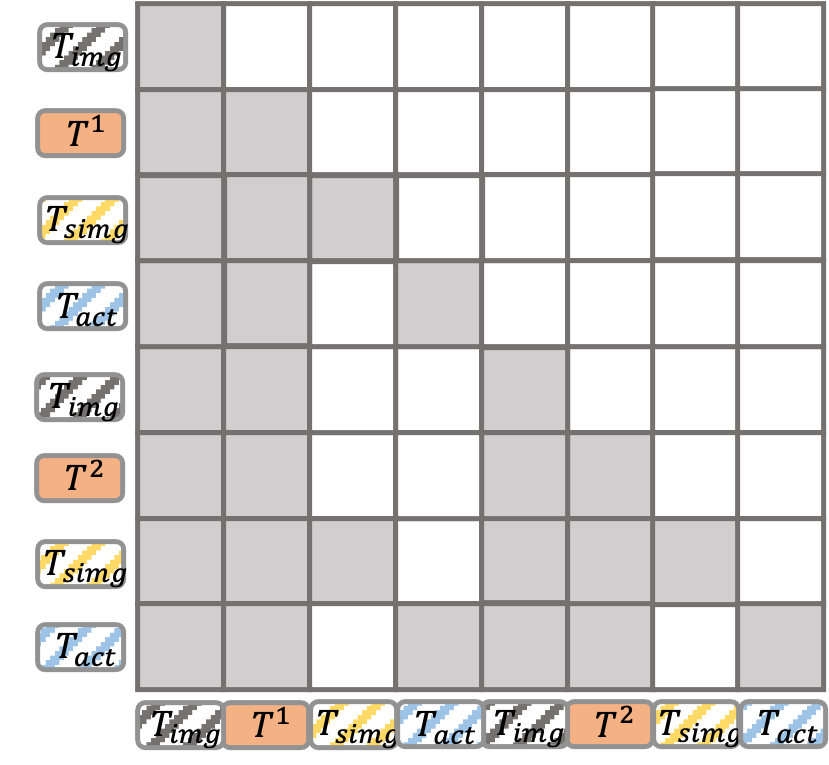}  
    \end{minipage}
    \hfill
    \begin{minipage}{0.24\textwidth}
        \centering
        \small
        \begin{tikzpicture}
        \begin{axis}[
            ybar,
            symbolic x coords={place, pick, turn, stack, push, grasp, close, put, open, rotate},
            xtick=data,
            x tick label style={rotate=45, anchor=east},
            xlabel style={at={(axis description cs:0.5,-0.1)}, anchor=north}, 
            ymin=0,
            ymax=4500,
            ytick={0, 2000, 4000},
            yticklabels={0, 2K, 4K},
            bar width=0.21cm,
            enlarge x limits=0.1,
            enlarge y limits=0.1,
            grid=major,
            width=1.2\textwidth,  
            height=3.9cm         
        ]
        \addplot coordinates {(place, 4335) (pick, 3494) (turn, 3452) (stack, 1654) (push, 1503) (grasp, 1196) (close, 1170) (put, 1091) (open, 688) (rotate, 583)};
        \end{axis}
        \end{tikzpicture}
    \end{minipage}
    \vspace{-2.5mm}
    \caption{\textbf{Left:} Modality Isolation Mask (MIM). The KQ mask structure regulates attention interactions among different modalities (e.g., \texttt{<text>}, \texttt{<image>}, \texttt{<action>}). Dark squares indicate allowed attention connections between keys (K) and queries (Q), while white squares denote prohibited attention, ensuring modality isolation.
    \textbf{Right:} Frequency of Tasks. This section illustrates the distribution of tasks within the dataset, detailing the number of episodes associated with each task. The bars represent the frequency of various tasks, including "place," "pick," and "turn," highlighting the diversity and focus areas of the dataset. The y-axis indicates the number of episodes, emphasizing the relative frequency of each task.} 
    \label{fig::others}
    \vspace{-5.5mm}
\end{figure}

(b) Attention Fusion: Continuing with the use of cross-attention in OpenFlamingo~\cite{awadalla2023openflamingo}, we fuse visual and text features wherein the text features \(F^h_T\) serve as the query, and the visual features \(U_I^h\) serve as the key and value. It is worth noting that the self-attention layer incorporates MIM (Figure~\ref{fig::others} Left), which allows training with auxiliary modality supervision and omitting unnecessary modalities during inference, significantly increasing the flexibility of modality.

(c) Multimodal Output Features: According to the modality read-out tokens, their corresponding output features are indexed as $F^h_{\text{simg}}$, $F^h_{\text{gimg}}$, $F^h_{\text{occ}}$, and \(F^h_{\text{act}}\).

\textbf{Multimodal Decoders:}
We design different decoder modules to accommodate various modalities.


(a) Image Decoder. As shown in Figure~\ref{fig:sub_imgd}, we design a simple structure that includes 2 attention decoder layers. This structure outputs image patches, which are then assembled into a complete image (static images $O^h_{simg}$ or wrist image $O^h_{gimg}$) based on their coordinates.

(b) Occupancy Decoder. The initial part of this structure, as shown in Figure~\ref{fig:sub_occd}, is similar to Image Decoder, generating the feature \( U^h_{occ} \). Then, \( U^h_{occ} \) is reshaped, upsampled, and processed through 3D convolutions to reconstruct the entire 3D occupancy $O^h_o= \{o^h_{pos}, o^h_{rgb}\}$.
The flexible model architecture allows the visual module to generate the occupancy map \( O^h_o = O^h_{o_v} \) from multi-view features using UVFormer, while LLM also outputs \( O^h_o = O^h_{o_m} \), corresponding to \( T_{occ} \). Experimental validation shows that \( O^h_{o_v} \) and \( O^h_{o_m} \) provide similar assistance for robotic manipulation. Unless otherwise specified, in this paper, \( O^h_o = O^h_{o_v} \).

(c) Action Decoder. In Figure~\ref{fig:sub_actd}, we use a few MLP layers or DiT~\cite{li2024cogact} blocks to output actions $O^h_A$, consisting of delta 6D poses $a_{pose}^{h}$ = \{$\Delta\text{pos}_x^h, \Delta\text{pos}_y^h, \Delta\text{pos}_z^h, \Delta\text{rot}_x^h, \Delta\text{rot}_y^h, \Delta\text{rot}_z^h\}$ and 1-DoF gripper actions $a_{g}^{h}$. Unless specified otherwise, this paper defaults to using an MLP action head. For more details on DiT, please refer to CogACT~\cite{li2024cogact}.

\subsection{Training Objective}

In multimodal learning tasks, we design the comprehensive loss function \(l\) to enhance overall model performance by optimizing different modality outputs:
\begin{align}
l = l_a + \lambda_{\text{image}} \left(l_{simg} + l_{gimg} \right) + \lambda_{\text{occ}} l_o.
\end{align}
Here, \(\lambda_{\text{image}}\) and \(\lambda_{\text{occ}}\) are balancing weights for image and occupancy losses. Notably, \(l_{simg}\), \(l_{gimg}\), and \(l_o\) can be excluded from training if the corresponding modality is unavailable, providing a flexible optimization framework.

\textbf{Action Loss \(l_a\).}
The action loss function optimizes \(a_{pose}^{h}\) and \(a_{gripper}^{h}\) using the combination of Mean Squared Error (MSE) and Binary Cross-Entropy (BCE):
\begin{align}
l_a = \sum_h \left(\mathrm{MSE}(a_{pose}^{h}, \hat{a}_{pose}^{h}) + \lambda_{g}\mathrm{BCE}(a_{g}^{h}, \hat{a}_{g}^{h}) \right).
\end{align}
Here, \(\lambda_{g}\) is a parameter used to adjust the weight of the gripper state loss, and \(\hat{a}_{pose}^{h}\) and \(\hat{a}_{g}^{h}\) represent the example action pose and gripper state at time step \(h\), respectively.

\textbf{Image Loss \(l_{simg}\) or \(l_{gimg}\). }
The image loss measures pixel-level differences  using L2 loss between the predicted images and the next frame images  \(\hat{I}^{h+1}_{simg}\) or \(\hat{I}^{h+1}_{gimg}\):
\begin{align}
l_{simg} &= \sum_h \sum_{\text{pixels}} \| O^h_{simg} - \hat{I}^{h+1}_{simg} \|_2^2, \\
l_{gimg} &= \sum_h \sum_{\text{pixels}} \| O^h_{gimg} - \hat{I}^{h+1}_{gimg} \|_2^2.
\end{align}

\textbf{Occupancy Loss \(l_o\): }
The occupancy loss incorporates spatial position and RGB color information:
\begin{align}
    l_o &= \sum_h \sum_{\text{points}} l_{o'}, \\
    l_{o'} = \| o^h_{pos} - \hat{o}^h_{pos} &\|_2^2 + \lambda_{rgb} \| o^h_{rgb} - \hat{o}^h_{rgb} \|_2^2,
\end{align}
where \(\lambda_{rgb}\) adjusts the contributions of position and RGB color losses.

\section{\textit{RoboData}}

The rise of ChatGPT~\cite{openai_chatgpt} and large AI models~\cite{touvron2023llama, alayrac2022flamingo, qwen} signifies a paradigm revolution in artificial intelligence, all built upon the foundation of rich ``internet-scale'' datasets. However, in the domain of embodied intelligence, research still focuses on single, specific tasks such as grasping, path planning, and pick-and-place, aiming to train agents tailored for particular scenarios. Although projects like Open X-Embodiment~\cite{padalkar2023open} and ARIO~\cite{wang2024robotsonenewstandard} compile multiple datasets, they still present numerous issues. For example, they lack essential 3D information—such as multi-view, camera intrinsic and extrinsic parameters, and depth maps—making these datasets suitable only for 2D multi-modal training. Moreover, there is a lack of proper spatial alignment across datasets; specifically, the 6D poses (i.e., position and orientation) of the robotic end-effectors recorded exhibit inconsistencies due to different world coordinate systems.

To address these challenges, we curate well-known publicly available datasets, including CALVIN~\cite{mees2022calvin}, Meta-World~\cite{yu2020meta}, LIBERO~\cite{liu2024libero}, Robomimic~\cite{robomimic2021}, RoboCasa~\cite{nasiriany2024robocasa}, ManiSkill2~\cite{gu2023maniskill2}, RoboCAS~\cite{zheng2024robocas}, RLBench~\cite{james2020rlbench}, Colosseum~\cite{pumacay2024colosseum} and RT-1~\cite{brohan2022rt}, forming a comprehensive dataset we call \textit{RoboData}, aiming to provide the industry with a complete and fair evaluation system. Specifically, \textit{RoboData} comprises 70,000 episodes and 7 million samples, and encompasses a diverse range of tasks, including placing, picking, turning, and stacking. Figure~\ref{fig::others} Right illustrates these tasks along with their corresponding number of episodes, highlighting the distribution of tasks within the dataset.

\begin{figure}
    \centering
    \begin{minipage}{0.22\textwidth}
        \centering
        \includegraphics[width=\textwidth]{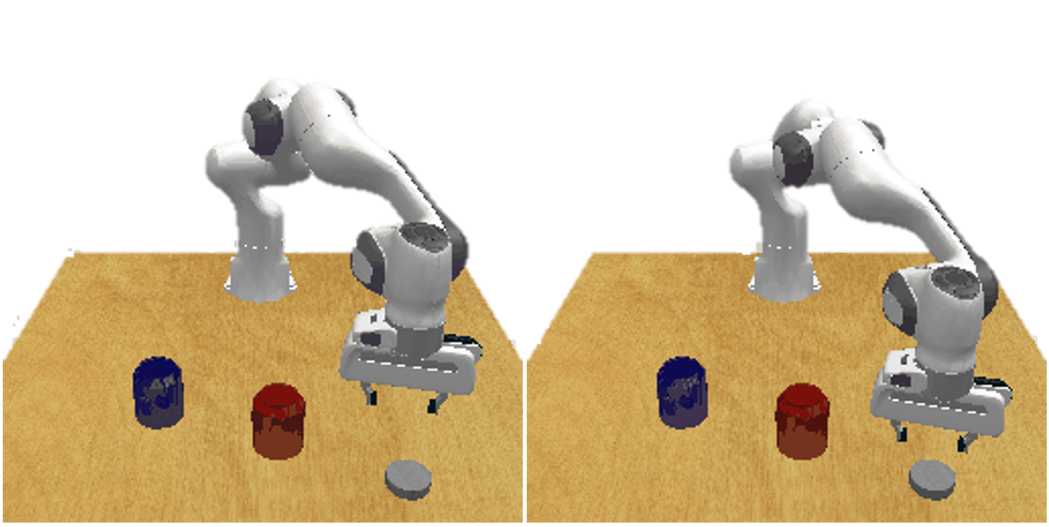}  
    \end{minipage}
    \hfill
    \begin{minipage}{0.22\textwidth}
        \centering
        \includegraphics[width=\textwidth]{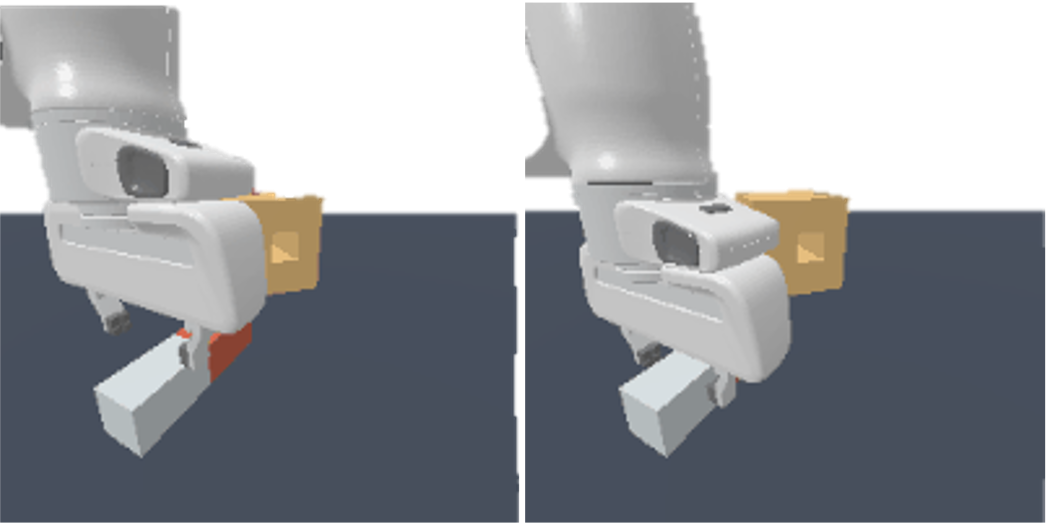}  
    \end{minipage}
    \vspace{-2.5mm}
    \caption{The downward movement of the robot in the RLBench and ManiSkill2 environments. }
    \label{fig::move_downward}
    \vspace{-6.5mm}
\end{figure}


As shown in Table~\ref{tab:robodata}, each dataset encompasses different simulation platforms and possesses unique world coordinate systems, workspaces, perspectives, and other characteristics. Therefore, we align the input and output spaces of the models based on multiple influencing factors.

\textbf{3D Space Alignment.} We focus on the unification of world coordinate systems, workspaces, and action spaces. Different datasets adopt their own coordinate systems. For example, the RLBench~\cite{james2020rlbench} and Colosseum~\cite{pumacay2024colosseum} reference the robot's body, setting the X-axis to point forward, the Y-axis to point left, and the Z-axis to point upward; whereas ManiSkill2~\cite{gu2023maniskill2} orient the X-axis forward, the Y-axis right, and the Z-axis downward. Figure~\ref{fig::move_downward} illustrate the movements of robotic arms from RLBench~\cite{james2020rlbench} and ManiSkill2~\cite{gu2023maniskill2} datasets, respectively. Although both exhibit similar motion directions (moving from top to bottom), the representation of actions differs significantly due to the variations in coordinate systems. For instance, in RLBench~\cite{james2020rlbench}, $a_{pose} = [0.0, 0.0, -0.1, 0.0, 0.0, 0.0]$, while in ManiSkill2~\cite{gu2023maniskill2}, $a_{pose} = [0.0, 0.0, 0.1, 0.0, 0.0, 0.0]$.

Unifying all data into the same coordinate system is crucial for conducting cross-platform joint training. If this unification is not achieved, conflicts may arise during the training process, negatively impacting the final learning outcomes. Benefiting from the supplementation of 3D information in the \textit{RoboData} dataset, we rotate the coordinate systems to unify all data to the same orientation: the X-axis pointing right, the Y-axis pointing forward, and the Z-axis pointing upward. For example, the original world coordinate system of  Robomimic~\cite{robomimic2021}, \(W_{ Robomimic}^{ori}\), is transformed into \(W_{ Robomimic}\):
$$
W_{ Robomimic} = \left[\begin{array}{cccc}
0 & 1 & 0 & 0 \\
-1 & 0 & 0 & 0 \\
0 & 0 & 1 & 0.4 \\
0 & 0 & 0 & 1 
\end{array}\right] W_{ Robomimic}^{ori}.
$$
In parallel, we limit the workspace through translation to the range of [-0.5, -0.5, 0] to [0.5, 0.5, 1].

\textbf{Action Representation Alignment.}
Various datasets employ different methods to obtain robot actions, and such diversities lead to inconsistencies in the data. For instance, CALVIN~\cite{mees2022calvin} uses the Euler Angle Difference Method (EADM) for action representation, while LIBERO~\cite{liu2024libero}, Robomimic~\cite{robomimic2021}, RoboCasa~\cite{nasiriany2024robocasa} utilize the Composite Rotation Matrix Method (CRMM), and  ManiSkill2~\cite{gu2023maniskill2} adopts the Pose Composition Method (PCM). To address this issue, we unify the action representation across all datasets by regenerating them using the CRMM, which has been validated in SRT~\cite{kim2024surgical}. This choice not only enhances data consistency but also provides a more reliable foundation for subsequent research. For more detailed technical information, 
please refer to Chapter~\ref{sec:action_repre}.

\textbf{Missing Data Imputation.}
In many datasets, camera intrinsics and extrinsics are often not directly provided, which poses challenges for research. To tackle this problem, we reconstruct the original simulations and render them anew, leveraging  original data to drive the robots in order to obtain these critical parameters. This process ensures data completeness and provides essential support for  experiments.

In summary, by effectively aligning the input and output spaces, we propose a new dataset standard aimed at comprehensively optimizing existing datasets. The implementation of this standard will facilitate the development of more versatile and general-purpose embodied AI agents. We plan to release this dataset along with evaluation code to promote further research and advancement in the field of embodied intelligence. Additionally, we welcome all researchers to join our efforts in driving the development of this field.

\begin{table}[ht]
    \small
    \centering
    \begin{tabular}{@{}>{\raggedright}m{2.1cm} @{} l @{\hspace{10pt}} m{2.1cm} @{} c @{}}
        \toprule
        \textbf{Dataset} & \textbf{Model} & \textbf{Source} & \textbf{SR$\uparrow$} \\
        \midrule
        \multirow{13}{*}{LIBERO} 
        & QueST~\cite{mete2024questselfsupervisedskillabstractions} & arXiv'24  & \underline{89.8\%} \\
        & VQ-BeT~\cite{lee2024behavior}             & ICML'24   & 81.4\% \\
        & MDT~\cite{reuss2024multimodal}            & RSS'24    & 67.2\% \\
        & MaIL~\cite{jia2024mail}                   & CoRL'24   & 60.3\% \\
        & PRISE~\cite{zheng2024prise}               & ICML'24   & 54.4\% \\
        & ATM~\cite{wen2023atm}                     & RSS'24    & 48.4\% \\
        & MUTEX~\cite{shah2023mutex}                & CoRL'23   & 53.0\% \\
        & DiffusionPolicy~\cite{chi2023diffusion}   & IJRR'23   & 75.4\% \\
        & ACT~\cite{zhao2023learning}               & RSS'23    & 46.6\% \\
        & ResNet-T~\cite{liu2024libero}             & NeurIPS'23 & 84.4\% \\
        & Distill-D~\cite{ha2023scaling}            & CoRL'23   & 49.9\% \\
        & RoboFlamingo$^{*}$~\cite{li2023vision}    & arXiv'23  & 72.1\% \\
        & $\pi0^{*}$~\cite{black2410pi0}            & arXiv'24  & 87.3\% \\
        & $\pi$-Fast$^*$~\cite{pertsch2025fast}       & arXiv'25  & 89.1\% \\
        & \iRoboMM (ours)                    & -         & \textbf{91.7\%} \\
        \midrule
        \multirow{3}{*}{RoboCasa} 
        & RoboCasa~\cite{robocasa2024}              & RSS'24    & \underline{28.8\%} \\
        & RoboFlamingo$^{*}$~\cite{li2023vision}    & arXiv'23  & 15.6\% \\
        & GR00T-N1-2B+IDM~\cite{bjorck2025gr00t}    & arXiv'25  & 40.9\% \\
        & \iRoboMM (ours)                    & -         & \textbf{47.4\%} \\
        \midrule
        \multirow{11}{*}{CALVIN} 
        & DTP~\cite{fan2025diffusion}               & arXiv'25  & 92.4\% \\
        & MDT~\cite{reuss2024multimodal}            & RSS'24    & \underline{93.7\%} \\
        & HULC++~\cite{mees23hulc2}                 & ICRA'24   & 93.0\% \\
        & SPIL~\cite{zhou2024languageconditioned}   & RA-L'24   & 84.6\% \\
        & LCD~\cite{zhang2022language}              & arXiv'23  & 88.7\% \\
        & RoboFlamingo~\cite{li2023vision}          & arXiv'23  & 86.0\% \\
        & PlayFusion~\cite{chen2023playfusion}      & CoRL'23   & 45.2\% \\
        & Distill-D~\cite{ha2023scaling}            & CoRL'23   & 86.7\% \\
        & HULC~\cite{mees2022hulc}                  & RA-L'22   & 82.7\% \\
        & CALVIN~\cite{mees2022calvin}              & RA-L'22   & 76.4\% \\
        & RoboFlamingo$^{*}$~\cite{li2023vision}    & arXiv'23  & 43.8\% \\
        & \iRoboMM (ours)                    & -         & \textbf{93.8\%} \\
        \midrule
        \multirow{8}{*}{Meta-World} 
        & PRISE~\cite{zheng2024prise}               & ICML'24   & \textbf{80.4\%} \\
        & PAD~\cite{guo2024prediction}              & NeurIPS'24 & 72.5\% \\
        & GR-1~\cite{wu2023unleashing}              & ICLR'24   & 57.4\% \\
        & SuSIE~\cite{black2023zero}                & ICLR'24   & 41.0\% \\
        & RT-2*~\cite{brohan2023rt}                 & arXiv'23  & 52.2\% \\
        & RT-1~\cite{brohan2022rt}                  & RSS'23    & 34.6\% \\
        & RoboFlamingo$^{*}$~\cite{li2023vision}    & arXiv'23  & 65.3\% \\
        & \iRoboMM (ours)                    & -         & \underline{80.1\%} \\
        \midrule
        \multirow{6}{*}{RT-1 Dataset} 
        & \textcolor{gray}{RT-2-X(55B)~\cite{padalkar2023open}} & \textcolor{gray}{ICRA'24} & \textcolor{gray}{60.7\%} \\
        & RT-1-X~\cite{padalkar2023open}            & ICRA'24   & \underline{53.4\%} \\
        & Octo-Base~\cite{team2024octo}             & RSS'24    & 16.9\% \\
        & OpenVLA~\cite{kim2024openvla}             & arXiv'24  & 24.8\% \\
        & HPT~\cite{wang2024scaling}                & NeurIPS'24& 35.0\% \\
        & \iRoboMM (ours)                    & -         & \textbf{60.0\%} \\
        \bottomrule
    \end{tabular}
    \vspace{-2.5mm}
    \caption{Performance on Various Datasets. \textbf{Bold indicates the best result}, \underline{underline indicates the second-best result}, and $^{*}$ denotes reproduced results, where RoboFlamingo$^{*}$ is the result of joint training on four simulation datasets.}
    \label{tab:sota}
    \vspace{-8.5mm}
\end{table}

\section{Experiments}

In the previous sections, we elaborate on the \iRoboMM{} framework and the characteristics of the \textit{RoboData} dataset. Next, we address the following questions using task success rates: its performance across multiple datasets, the importance of each module within \iRoboMM{} and \textit{RoboData}.

\subsection{Results and Analysis}
\label{sec:sota}

\begin{table*}[!ht]
    \small
    \centering
    \addtolength{\tabcolsep}{6pt}
    \begin{tabular}{@{}c|cccc|cccccc@{}} 
        \toprule
        \multirow{2}{*}{ID}  & \multirow{2}{*}{FFA} & \multirow{2}{*}{Image} & \multirow{2}{*}{UVFormer} & \multirow{2}{*}{OCC} & \multicolumn{5}{c}{Task Completed in a Sequence } &   \multirow{2}{*}{Avg Len} \\
        \cline{6-10} 
            &           &           &               &       & 1  &   2 &   3 &   4 &   5 &  \\
        \midrule
        1   &\xmarkg & \xmarkg & \xmarkg & \xmarkg & 81.0\% & 48.1\% & 25.7\% & 14.5\% & 8.6\% & 1.77 \\
        2   &\cmarkg & \xmarkg & \xmarkg & \xmarkg & 85.0\% & 63.3\% & 42.0\% & 28.7\% & 18.8\% & 2.37 \\
        3   &\cmarkg & \cmarkg & \xmarkg & \xmarkg & 88.5\% & 74.7\% & 60.7\% & 49.1\% & 39.6\% & 3.13 \\
        4   &\cmarkg & \xmarkg & \cmarkg & \xmarkg & 94.2\% & 74.7\% & 55.1\% & 38.3\% & 25.8\% & 2.88 \\
        5   &\cmarkg & \cmarkg & \cmarkg & \xmarkg & 94.1\% & 78.9\% & 63.7\% & 48.0\% & 36.4\% & 3.21 \\
        6   &\cmarkg & \xmarkg & \cmarkg & \cmarkg & 94.5\% & 78.4\% & 61.1\% & 46.6\% & 35.4\% & 3.18 \\
        \midrule
        7   &\cmarkg & \cmarkg & \cmarkg & \cmarkg & 94.7\% & 80.3\% & 65.1\% & 51.4\% & 39.0\% & 3.31 \\   
        7*  &\cmarkg & \cmarkg & \cmarkg & \cmarkg & 96.9\% & 83.0\% & 68.1\% & 56.5\% & 46.8\% & 3.51 \\   
        \bottomrule
    \end{tabular}
    \vspace{-2.5mm}
    \caption{Performance on different modules of \iRoboMM. FFA and OCC refer to frame by frame action module and occupancy supervision. Row 7* signifies the replacement of the MLP action head with the DiT action head. }
    \label{tab::ablation}
    \vspace{-4mm}
\end{table*}

\begin{figure*}[ht]
    \centering
    \includegraphics[width=0.8\textwidth]{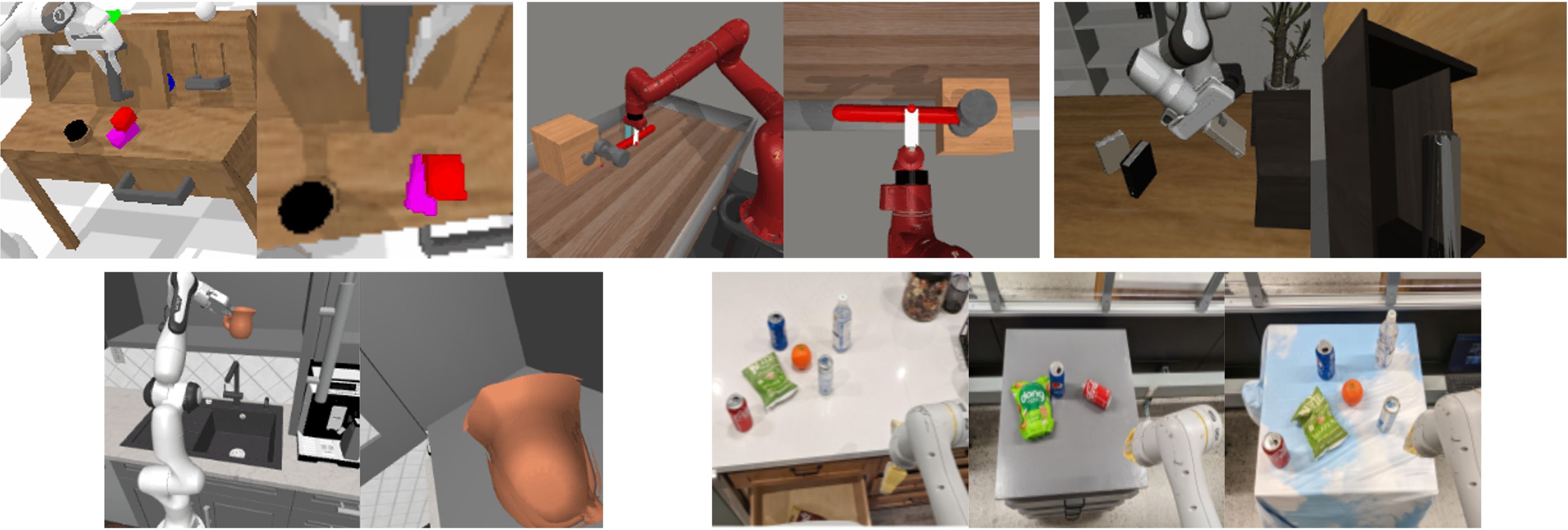}  
    \caption{Evaluation Datasets. We evaluate \iRoboMM{} across five simulation benchmarks and present policy rollout visualizations of the experiments. From left to right, the benchmarks include CALVIN, Meta-World, LIBERO, RoboCasa, and  Robomimic. Experiment details can be found in Section~\ref{sec:sota}.}
    \label{fig::robodata_images}
\end{figure*}

\textbf{Protocol.} 
Due to the substantial time and computational resources required for model training, this study conducts joint training of the \iRoboMM{} model with DiT action head across five datasets. These datasets encompass simulated data (CALVIN~\cite{mees2022calvin}, Meta-World~\cite{yu2020meta}, LIBERO~\cite{liu2024libero}, RoboCasa~\cite{nasiriany2024robocasa}) and RT-1~\cite{brohan2022rt}, as shown in the Figure~\ref{fig::robodata_images}. The \iRoboMM{} model, with 4 billion parameters in bf16 precision, is trained for 10 epochs using 2.1 million samples on 32 80G-A100 GPUs, taking about 50 hours.
During the evaluation phase, we adhere to the official configurations provided for each dataset. 
The evaluation metric adopted is success rate(SR), to comprehensively assess the model's performance.

In this study, we are conducting systematic experiments to evaluate the performance of the \iRoboMM{} model in multi-data fusion training and its comparison with expert models. As shown in Table~\ref{tab:sota}, \iRoboMM{} demonstrates impressive performance across several simulation datasets. In the LIBERO~\cite{liu2024libero} dataset, \iRoboMM{} achieves a success rate of 91.7\%, significantly surpassing the current best model, QueST~\cite{mete2024questselfsupervisedskillabstractions}, which has an 89.8\% success rate. In the RoboCasa~\cite{nasiriany2024robocasa} dataset, the model achieves a success rate of 47.4\%, showcasing its robust adaptability in diverse scenarios. For the CALVIN~\cite{mees2022calvin} and Meta-World~\cite{yu2020meta} datasets, the success rates are 93.8\% and 80.1\%, respectively, comparable to other leading expert models such as the MDT~\cite{reuss2024multimodal} and the PRISE~\cite{zheng2024prise} model. This demonstrates its balanced capabilities in temporal reasoning and multi-task learning.

Notably, \iRoboMM{} outperforms the general model RoboFlamingo$^{*}$~\cite{li2023vision} across all four datasets. For instance, there is a 19.6\% improvement in success rate in the LIBERO~\cite{liu2024libero} dataset and a 14.8\% increase in the Meta-World~\cite{yu2020meta} dataset. This significant performance gap not only highlights the architectural advantages and innovation of \iRoboMM{} but also reflects the potential and superiority of its spatial awareness and multimodal integration strategies in executing complex tasks.

Specifically, as shown in Table~\ref{tab:sota}, \iRoboMM{} achieves a 60\% average task success rate on the Google RT-1~\cite{brohan2022rt} benchmark dataset, significantly outperforming models with equivalent parameter sizes. The research focuses on extracting three complex operational tasks (Pick Coke Can, Move Near, Open/Close Drawer). Through structured reassembly and distribution alignment techniques, we construct the multimodal joint training framework, which combines with four \textit{RoboData} subsets. The system validation is completed under the SimpleEnv evaluation framework.
In terms of experimental design, to address the sensor parameter missing issues in the RT-1~\cite{brohan2022rt} dataset, we supplement them based on the SimpleEnv test environment. Given the inaccessibility of depth perception data, we omit occlusion-aware loss during training. Moreover, to handle the lack of wrist camera sensor data, we introduce a dynamic mask filling mechanism during the preprocessing stage.

It is important to note that, unlike the general representation learning paradigm adopted by \iRoboMM{} and RoboFlamingo$^{*}$, all compared models are expert models utilizing task-specific optimization strategies, which restrict parameter updates to a single data domain. Although existing studies include multi-dataset pretraining schemes, they typically rely on fine-tuning adaptations to the target dataset. This domain-dependent optimization approach can achieve local performance gains but also increases training complexity and significantly reduces the model's cross-domain generalization ability.



\subsection{Ablation Study}

\begin{figure}[t]
    \vspace{-2.5mm}
    \centering
    \includegraphics[width=1.0\linewidth]{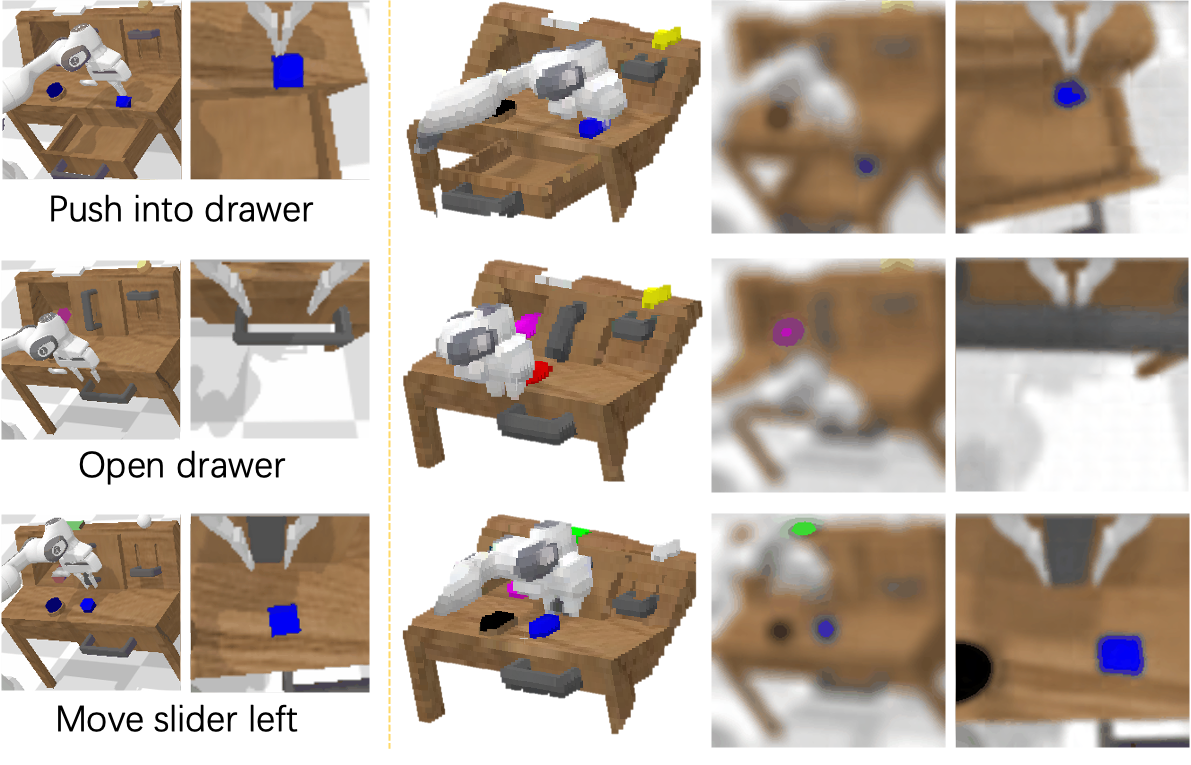} 
    \vspace{-7.5mm}
    \caption{The illustration demonstrating the auxiliary modality results of \iRoboMM{} on the Calvin three tasks. The inputs consist of static images, wrist-view images, and text, while the outputs include the current OCC and the predicted next frame's static and wrist-view images.}
    \label{fig:output_calvin}
    \vspace{-6mm}
\end{figure}
\subsubsection{\iRoboMM{} Module}
\textbf{Protocol.} 
This experiment aims to systematically evaluate the contributions of each module in the \iRoboMM{} framework to overall performance, using the CALVIN~\cite{mees2022calvin} dataset for analysis. We design seven different experimental setups to progressively analyze the impact of each module. First, in the baseline setup, the model receives $H$ frames as input and outputs an action only at the last frame to assess its basic prediction capability. Next, with the addition of the Frame by Frame Action~(FFA) module, the model is able to output actions after each frame, allowing us to examine its immediate prediction capability and performance in continuous prediction scenarios.
As the experiment progresses, the introduction of the Image module enables the model to generate the next frame, thereby enhancing its contextual understanding and fine-grained perception. Subsequently, we added the UVFormer module to assess its specific contribution to the model's performance. Finally, by incorporating the occupancy (OCC) output, we further enhance the model's understanding of 3D spatial information, thereby improving its predictive capabilities.

\textbf{Experiment.} 
The results of the experiments are summarized in Table~\ref{tab::ablation}, which demonstrates a clear trend of performance improvement as different modules are added. The baseline configuration achieves a success rate of 81.0\% for completing the first task, while the addition of the FFA module increases this success rate to 85.0\%. The introduction of image output capability further enhances success rates across all tasks. Notably, the UVFormer setup shows a significant improvement in the first task, achieving a success rate of 94.2\%, indicating the effectiveness of the UVFormer module in enhancing the model's predictive capabilities. The OCC configuration builds on this by incorporating occupancy loss, resulting in even higher success rates, particularly in later tasks, suggesting that the model benefits from enriched contextual and spatial information.

Finally, row 7 shows the highest performance across all tasks, with a significant increase in the average length of completed tasks. Combining insights from Table~\ref{tab::ablation} and Figure~\ref{fig:output_calvin}, it is evident that although the generated images and occupancy do not exhibit ideal quality, they significantly contribute to enhancing action performance. This indicates that even with suboptimal generation quality, the integration of visual and spatial information still has a positive impact on the execution of robotic tasks.

Overall, this ablation study highlights the contributions of each module within \iRoboMM, illustrating how enhancements can lead to significant performance gains.

\subsubsection{RoboData Alignment}

The training of RoboFlamingo and \iRoboMM{} is conducted using both non-aligned and aligned data across the four datasets: LIBERO, RoboCasa, CALVIN, and Meta-World. As indicated in the table, due to RoboFlamingo's two-dimensional feature input, it cannot achieve input space alignment, resulting in performance that does not meet expectations even when trained on aligned \textit{RoboData}.
The results of \iRoboMM$^{-}$ reflect the effects of training on non-aligned \textit{RoboData}. The research indicates that misalignment between input and output spaces significantly hinders model performance when training across multiple datasets. Notably, in the Meta-World dataset~\cite{yu2020meta}, the impact of this misalignment is relatively minimal. This is because the operations are limited to changes in three positions without rotational variations ($a_{pose}^{h} = \{\Delta\text{pos}_x^h, \Delta\text{pos}_y^h, \Delta\text{pos}_z^h, 0.0, 0.0, 0.0\}$), and the original spatial coordinate system is consistent with that of the \textit{RoboData}. Therefore, no additional alignment process is required, and model performs relatively well even when not trained on aligned data.
{The successful alignment of input and output spaces within \textit{RoboData}, combined with the compatible architecture of \iRoboMM, enables effective joint training and evaluation across multiple datasets.}

\begin{table}[t]
    \vspace{-2.5mm}
    \centering
    \small
    \begin{tabular}{@{}l@{\hspace{5pt}} c@{\hspace{5pt}}c@{\hspace{5pt}}c@{\hspace{5pt}}c@{}}
        \toprule
        Model               & LIBERO & RoboCasa & CALVIN & Meta-World   \\
        \midrule
        RoboFlamingo$^{*-}$ & 69.1\% & 15.8\%   & 41.7\% & 63.2\%       \\
        RoboFlamingo$^{*}$  & 72.1\% & 15.6\%   & 43.8\% & 65.3\%       \\
        \iRoboMM$^{-}$        & 64.2\% & 27.0\%   & 74.7\% & 79.3\%       \\
        \iRoboMM              & 90.7\% & 30.6\%   & 91.0\% & 78.6\%       \\
        \bottomrule
    \end{tabular}
    \vspace{-3.0mm}
    \caption{The results of RoboFlamingo and \iRoboMM{} before and after data alignment.}
    \label{tab:sota_with_roboflamingo}
    \vspace{-4.5mm}
\end{table}

\vspace{-2.5mm}
\section{Conclusion and Outlook}



Enhancing large model performance relies on effectively utilizing vast, consistently formatted data. In the EAI field, robotic operation data is limited. A feasible approach is for institutions to collect data using their robotic devices and consolidate it into a large, heterogeneous dataset to drive industry development. Efficiently using this data is challenging. In RT-X Table 1, direct data fusion results in RT-1-X underperforming compared to RT-1, highlighting the difficulty in utilizing heterogeneous data effectively.

This paper introduces the method to effectively utilize heterogeneous data from various institutions through the model architecture featuring 3D perception input and multimodal output, combined with data reorganization. \iRoboMM{} and \textit{RoboData} synergize effectively. For instance, capturing the same scene with cameras of different parameters can result in varying 2D visual features, impacting action outputs. In contrast, the 3D scene remains consistent, and UVFormer's 3D features exhibit minimal variation. Without \iRoboMM's 3D spatial input, achieving input space alignment is impossible.
We initially validate this approach using simulations, leveraging the accessibility of heterogeneous data in simulators. 

Although our training objectives in supervised learning have reached only a moderate scale, \iRoboMM{} and \textit{RoboData} open new avenues for cross-embodiment joint training and evaluation. We hope this work provides the industry with a comprehensive and fair evaluation system, inspires future exploration in foundational robotic model research, and enhances the generality and performance of robotic learning overall.

\newpage

{
    \small
    \bibliographystyle{ieeenat_fullname}
    \bibliography{main}
}


\clearpage
\setcounter{page}{1}
\maketitlesupplementary

\section{Related Work}

\textbf{Robotic Datasets.} In the early stages of robotics research, it is typically necessary to collect specific datasets for each robot, task, and environment, such as RLBench~\cite{james2020rlbench} and CALVIN~\cite{mees2022calvin}. Although these datasets are highly customized and of high quality, they are limited in quantity and have poor generalization capabilities. To further enhance model performance and generalization, researchers have collected large amounts of data through teleoperation methods, such as RT-1~\cite{brohan2022rt} and RH20T~\cite{fang2023rh20t}. These large-scale datasets cover more scenarios and tasks, supporting multi-task learning, but also bring high data annotation costs. As research progresses, methods for integrating multiple datasets, such as Open X-Embodiment~\cite{padalkar2023open} and DROID~\cite{khazatsky2024droid}, have been proposed to improve model generalization and data utilization efficiency by merging data from different sources. However, these methods also face issues of data inconsistency and potential biases. This paper proposes \textit{RoboData}, which efficiently integrates multiple datasets and unifies the input and output spaces, thereby addressing data heterogeneity. Additionally, it breaks the limitation of training for a single specific task, providing a unified benchmark for robotic manipulation.

\textbf{Robotic Policies.} Previous works such as R3M~\cite{nair2022r3m}, VC-1~\cite{majumdar2023we}, ACT~\cite{zhao2023learning}, and HULC++~\cite{mees2023grounding} typically employ strategies with a small number of parameters. Subsequent models like RoboFlamingo~\cite{li2023vision}, Corki~\cite{huang2024corkienablingrealtimeembodied}, and RoboUniView~\cite{liu2024robouniview} have built on multimodal large models but have only fine-tuned on limited datasets. Despite advancements in multi-task learning and few-shot learning, recent models such as RT-X~\cite{padalkar2023open}, Octo~\cite{team2024octo}, HPT~\cite{wang2024scaling}, CrossFormer~\cite{doshi24-crossformer}, GR-2~\cite{cheang2024gr2generativevideolanguageactionmodel}, and OpenVLA~\cite{kim2024openvla} have trained vision-language-action robotic policies on various datasets. However, these works often pre-train on data from real robots~\cite{fang2023rh20t, khazatsky2024droid}, human videos~\cite{nair2022r3m, grauman2022ego4d}, and simulation domains~\cite{zheng2024robocas, mees2022calvin}, neglecting the uniformity of physical space, and achieve good performance only after fine-tuning on specific datasets. Given that robots operate in 3D physical environments, their perception and interaction capabilities must integrate 3D sensing, akin to the requirements of autonomous driving systems.

\section{Real world experiment}
\begin{figure}[ht]
    \vspace{0mm}
    \centering 
    \includegraphics[width=1.0\linewidth]{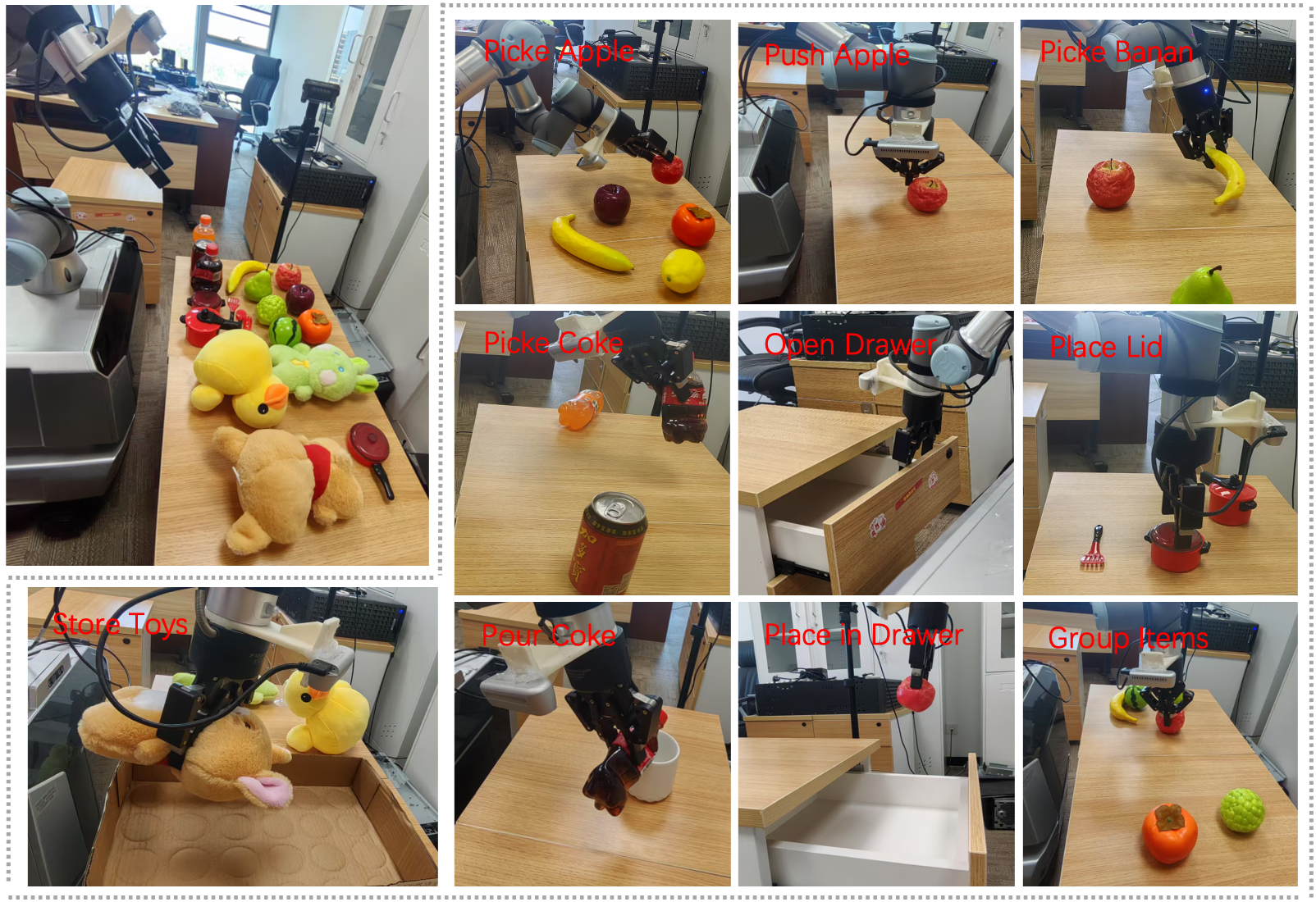} 
    \vspace{-7mm}
    \caption{Real-world setup with 10 tasks, featuring a fixed Dalu base, UR3 arm, Robotiq gripper, wrist-mounted Intel D435, and ground-mounted Orbbec Gemini Pro. Hand-eye calibration and point cloud fusion enable OCC-supervised training. The system runs on ROS1 with an NVIDIA RTX 3090 GPU.}
    \vspace{-4mm}
    \label{fig:setup}
\end{figure}

To evaluate \iRoboMM{} performance in real-world scenarios, we constructed a physical evaluation system as illustrated in Figure~\ref{fig:setup}. The robotic platform comprises a Dalu mobile base and an UR3 robotic arm. During experiments, the mobile base remains stationary, and only the robotic arm retaining degrees of freedom. The system is equipped with essential perception and actuation components, including a Robotiq two-finger gripper, an Intel D435 depth camera mounted on the wrist of the arm (denoted as $cam_{wrist}$), and an Orbbec Gemini Pro depth camera fixed to a stand on the ground to the left of the arm (denoted as $cam_{static}$).

Since \iRoboMM{} requires camera extrinsic parameters, we perform hand-eye calibration prior to experiments. Using the image data and intrinsic/extrinsic parameters from both $cam_{wrist}$ and $cam_{static}$, we constructed point clouds of the scene, which were used for occupancy (OCC) supervised training to enable the model to learn 3D geometric structures. The system operates on a ROS1-based communication framework to enable efficient interaction between the Nvidia 3090 server and the robotic arm.

In Figure~\ref{fig:setup}, we design ten real-world tasks grouped into three difficulty levels.
Easy: pick or push apple, pick banana or Coke bottle.
Medium: open drawer, place lid on pot, pour Coke into cup.
Hard: open drawer and place object inside, group similar items, store plush toys.
We collect 100 teleoperation demonstrations per task (1000 total) for training. For evaluation, each task undergoes 10 trials with manual resets and clear success criteria (e.g., lifting an apple at least 5 cm). More details are provided in the supplementary due to space limits.
We compare two baseline models: RoboFlamingo (2D) and RoboUniView (3D), which, like \iRoboMM, are both 3D models.

In Table~\ref{tab:combined} Right, \iRoboMM{} outperforms baselines across all difficulty levels, especially on medium and hard tasks, demonstrating stronger generalization and 3D reasoning. While RoboUniView performs reasonably on easy tasks, it struggles in more complex scenarios. RoboFlamingo, lacking explicit 3D scene modeling, shows weaker overall performance.

\begin{table}[ht]
    \centering
    \begin{subtable}[t]{0.45\textwidth}
        \centering
        \begin{tabular}{@{}l@{\hspace{1pt}}c@{\hspace{1pt}}c@{\hspace{1pt}}c@{}}
            \toprule
            Model            & Easy    & Medium  & Hard   \\
            \midrule
            \iRoboMM           & 82.5\%  & 70.0\%  & 46.6\% \\
            RoboUniView$^*$  & 75.0\%  & 36.6\%  & 23.3\% \\
            RoboFlamingo$^*$ & 35.0\%  & 6.6\%   & 6.6\%  \\
            \bottomrule
        \end{tabular}
    \end{subtable}
    \caption{Success rates by task difficulty, $^*$ indicates reproduced results.}
    \label{tab:combined}
    \vspace{-5mm}
\end{table}

\section{\iRoboMM{} detailed information}

In summary, the parameters used in the study are as follows: \( H = 12 \), \( N = 3 \), \( H = 256 \), \( W = 256 \), \( L = 80 \), \( B = 80 \), \( P = 40 \), \( C = 1024 \), \( \lambda_{\text{image}} = 0.1 \), \( \lambda_{\text{occ}} = 0.1 \), \( \lambda_{g} = 0.01 \), and \( \lambda_{\text{rgb}} = 0.5 \). The optimization strategy employs AdamW, while the learning rate schedule utilizes cosine annealing, with an initial learning rate of \( 10^{-4} \) and a termination rate of \( 10^{-6} \). The model is trained for a total of 10 epochs unless otherwise specified.

\section{Action Representation}
\label{sec:action_repre}

Different datasets have different methods for obtaining actions. For example, given the poses at two consecutive time steps  \( P^{t} = (p^{t}, r^{t}_{quat}) \) and \( P^{t+1} = (p^{t+1}, r^{t+1}_{quat}) \), which are represented by 3D coordinates and quaternions, respectively.

\subsection{Euler Angle Difference Method (EADM)}
The Euler Angle Difference Method is a way to describe rotational transformations by calculating the difference in Euler angles between two poses (or orientations). The specific steps are as follows:

1. Convert the quaternions \( r^{t}_{quat} \) and \( r^{t+1}_{quat} \) to Euler angles \( r^{t}_{euler} \) and \( r^{t+1}_{euler} \), respectively.

2. Compute the differences in the 3D coordinates and Euler angles to obtain the action:
\begin{align}
     A_t = (p^{t+1} - p^{t}, r^{t+1}_{euler} - r^{t}_{euler}).
\end{align}

This method is intuitive and easy to understand, but it may encounter gimbal lock issues when dealing with large-angle rotations or multiple rotations.

\subsection{Composite Rotation Matrix Method (CRMM)}
The Composite Rotation Matrix Method describes complex rotational transformations by multiplying multiple rotation matrices. A rotation matrix is a linear algebra tool used to represent rotations in three-dimensional space. The specific steps are as follows:

1. Convert the quaternions \( r^{t}_{quat} \) and \( r^{t+1}_{quat} \) to rotation matrices \( r^{t}_{matrix} \) and \( r^{t+1}_{matrix} \), respectively.

2. Compute the composite rotation by multiplying the rotation matrices to obtain the action:
\begin{align}
   A_t = (p^{t+1} - p^{t}, r^{t+1}_{matrix} \cdot Inv(r^{t}_{matrix}))
\end{align}

This method is advantageous because it can conveniently handle any complex combination of rotations and avoids the gimbal lock problem.

\subsection{Pose Composition Method (PCM)}
The pose composition method is a way to describe the position and orientation of an object in space. By combining the poses at two consecutive time steps, complex motions can be described. The specific steps are as follows:

1. Convert the quaternions \( r^{t}_{quat} \) and \( r^{t+1}_{quat} \) to rotation matrices \( r^{t}_{matrix} \) and \( r^{t+1}_{matrix} \), respectively.

2. Combine the poses to obtain the action:
\begin{align}
   A_t = \left(Inv(R^{t}_{matrix}) \cdot (p^{t+1} - p^{t}), Inv(R^{t}_{matrix}) \cdot R^{t+1}_{matrix}\right)
\end{align}

This method is advantageous because it can conveniently describe and compute complex motions of objects in space and is widely used in robotics and computer vision.

\begin{sidewaystable*}
\small 
\centering
\renewcommand{\arraystretch}{2} 

\begin{tabularx}{\textwidth}{ c c c  c  c  c  c  c  c }
\hline
\textbf{Platform} & \textbf{Physics Engine} & \textbf{Robot} & \textbf{\makecell{Coordinate \\(X-Y-Z)}} & \textbf{Views} & \textbf{\makecell{Camera \\Parameters}} & \textbf{\makecell{Action\footnote{EADM: Euler Angle Difference Method; CRMM: Composite Rotation Matrix Method; PCM:Pose Composition Method.} \\Representation}} & \textbf{Tasks} & \textbf{Episodes} \\
\hline
CALVIN~\cite{mees2022calvin}      & PyBullet                  & 7-DOF Franka  & Right-Forward-Up  & Static, Gripper                                                           & No    & EADM      & 34 & 20K \\
Meta-World~\cite{yu2020meta}   & MuJoCo                    & 4-DOF Sawyer  & Right-Forward-Up  & \makecell{behindGripper, corner, corner2,\\ corner3, topview, gripperPOV} & No    & None      & 50  & 5K \\
Libero~\cite{liu2024libero}      & MuJoCo                    & 7-DOF Franka  & Forward-Left-Up   & \makecell{frontview, birdview, \\agentview, sideview}                     & No    & CRMM      & 130 & 6.5K \\
RoboMimic~\cite{robomimic2021}   & MuJoCo                    & 7-DOF Franka  & Forward-Left-Up   & \makecell{agentview, robot0\_eye\_in\_hand}                               & No    & CRMM      & 8   & 1.6K \\
RoboCasa~\cite{robocasa2024}    & MuJoCo                    & 12-DOF Franka & Forward-Left-Up   & \makecell{center, left, \\ right, frontview, eye\_in\_hand}               & No    & CRMM      & 100 & 5K \\
ManiSkill2~\cite{gu2023maniskill2}  & SAPIEN                    & 7-DOF Franka  & Forward-Right-Down& \makecell{base\_camera, hand\_camera}                                     & No    & PCM       & 20  & 30K \\
RoboCAS~\cite{zheng2024robocas}     & \makecell{SAPIEN\\/Isaac} & 7-DOF Franka  & Forward-Left-Up   & \makecell{gripper\_camera, base\_camera, \\ static\_camera}               & Yes   & Absolute & 3   & 7.3K \\
RLBench~\cite{james2020rlbench}     & V-REP                     & 7-DOF Franka  & Forward-Left-Up & \makecell{left\_shoulder, right\_shoulder, \\ wrist, front}               & Yes   & Absolute & 18  & 1.8K \\
Colosseum~\cite{pumacay2024colosseum}   & PyRep                     & 7-DOF Franka  & Forward-Left-Up & \makecell{left\_shoulder, right\_shoulder, \\ wrist, front}               & Yes   & Absolute & 20  & 2K \\
\hline
\end{tabularx}

\begin{tabularx}{\textwidth}{ c@{\hspace{25pt}}  c@{\hspace{25pt}}  c@{\hspace{25pt}}  c }
\hline
\textbf{Platform} & \textbf{\makecell{Workspace \\ $\text{Min}_{\left[X, Y, Z\right]},\text{Max}_{\left[X, Y, Z\right]}$}} & \textbf{\makecell{Action Space \\$\text{Min}_{\left[X, Y, Z, Pitch, Roll, Yaw\right]},\text{Max}_{\left[X, Y, Z, Pitch, Roll, Yaw\right]}$}} & \textbf{\makecell{Gripper \\(Open/Close)}} \\
\hline
CALVIN~\cite{mees2022calvin} &  [-0.43, -0.57, 0.43], [0.37, -0.00, 0.80] & [-0.03, -0.03, -0.03, -6.28, -0.07, -6.27], [0.04, 0.02, 0.02, 6.28, 0.06, 6.28] & -1/1 \\
Meta-World~\cite{yu2020meta} & [-0.50, -0.10, 0.12], [0.48, 0.41, 0.60] & [-1.00, -1.00, -1.00], [1.00, 1.00, 1.00] & 0.5/-0.5 \\
Libero~\cite{liu2024libero} & [-0.24, -0.43, 0.01], [0.86, 0.57, 0.90] & [-0.93, -0.93, -0.93, -0.33, -0.37, -0.37], [0.93, 0.93, 0.93, 0.37, 0.37, 0.37] & 1/-1 \\
RoboMimic~\cite{robomimic2021} & [-0.17, -0.40, 0.90], [0.33, 0.33, 1.29] & [-1.0, -1.0, -1.0, -0.55, -1.0, -1.0], [1.0, 1.0, 1.0, 0.72, 0.45, 1.0] & 1/-1 \\
RoboCasa~\cite{robocasa2024} & [-0.81, -1.35, 0.70], [0.85, 0.75, 1.83] & [-1.0, -1.0, -1.0, -1.0, -1.0, -1.0], [1.0, 1.0, 1.0, 1.0, 1.0, 0.89] & 1/-1 \\
ManiSkill2~\cite{gu2023maniskill2} & [-0.26, -0.79, -1.17], [0.85, 0.76, 0.00] & [-0.14, -0.15, -0.16, -0.09, -0.09, -0.09,], [0.17, 0.16, 0.15, 0.09, 0.09, 0.09] & -1/1 \\
RoboCAS~\cite{zheng2024robocas} & [-0.70, -0.82, 0.062], [0.85, 0.67, 0.92] & [-0.04, -0.04, -0.04, -0.12, -0.10, 0.15], [0.03, 0.04, 0.03, 0.07, 0.09, 0.16, 0.08] & 0/0.08  \\
RLBench~\cite{james2020rlbench} & [-0.89, -0.72, 0.80], [0.56, 0.69, 1.89] & [-0.05, -0.04, -0.03, -1.0, -0.15, -1.0], [0.05, 0.06, 0.03, 1.0, 0.15, 1.0] & 0/1 \\
Colosseum~\cite{pumacay2024colosseum} & [-0.68, -0.68, 0.83], [0.54, 0.70, 1.85] & [-0.04, -0.04, -0.04, -0.79, -0.12, -0.76], [0.03, 0.03, 0.04, 0.79, 0.35, 0.79] & 0/1 \\
\hline
\end{tabularx}

\caption{Detailed information of CALVIN~\cite{mees2022calvin}, Meta-World~\cite{yu2020meta}, LIBERO~\cite{liu2024libero}, Robomimic~\cite{robomimic2021}, RoboCAS~\cite{zheng2024robocas}, ManiSkill2~\cite{gu2023maniskill2}, RoboCasa~\cite{nasiriany2024robocasa}, RLBench~\cite{james2020rlbench}, and Colosseum~\cite{pumacay2024colosseum}.}
\label{tab:robodata}
\end{sidewaystable*}

\section{\textit{RoboData} detailed information}

\subsection{CALVIN Dataset}
CALVIN is an open-source simulated benchmark specifically designed for learning long-horizon language-conditioned tasks in robotics. The dataset features four distinct environment splits, labeled A, B, C, and D. Each environment contains 6 hours of human-teleoperated recording data, resulting in over 2 million trajectories. However, only 1\% of this data is annotated with language instructions, amounting to approximately 24,000 trajectories. Each environment split is uniquely configured with various objects and scenarios, allowing for comprehensive validation of the performance, robustness, and generality of the trained policies across different data combinations.

The benchmark utilizes a 7-degree-of-freedom (7-DOF) Franka Emika Panda robotic arm equipped with a parallel gripper. This robotic platform is enhanced with onboard sensors and captures images from two camera perspectives, enabling it to effectively execute complex sequences of language instructions. The coordinate system is based on the robot's body, represented as Right-Forward-Up, where the X-axis represents the right direction, the Y-axis denotes the forward direction, and the Z-axis indicates the upward direction.

For action representation, CALVIN employs EADM. To ensure that actions are appropriately scaled for network predictions, specific scaling factors are applied: 0.02 for the X, Y, and Z axes, and 0.05 for the pitch, roll, and yaw angles. The states of the gripper are represented using -1 for open and 1 for closed, facilitating clear action commands.

\textbf{Space Alignment:} \textit{RoboData} includes all 34 distinct tasks, providing 20,000 episodes with language instructions for training. Action representations are regenerated using CRMM, and camera parameters are obtained through replay. Since the other input spaces are consistent with those predefined by \textit{RoboData}, no alignment adjustments are necessary.

The dataset evaluates 1,000 unique instruction chains, focusing primarily on sequential task execution. In each experiment, the robotic agent successfully completes a series of up to five language instructions in succession. The agent can only proceed to the next instruction after successfully achieving the current task, establishing a clear dependency on the completion of prior actions.

\subsection{Meta-World Dataset}

Meta-World is a tabletop manipulation benchmark designed to facilitate the training and evaluation of robotic manipulation policies in a simulated environment. This dataset focuses on the reinforcement learning domain and does not release training data. The simulator includes six perspectives: behindGripper, corner, corner2, corner3, topview, gripperPOV.

The benchmark utilizes a 4-degree-of-freedom (4-DOF) Franka Emika Panda robotic arm equipped with a parallel gripper, which does not allow end rotation. The gripper states are represented by the numbers 0.5/-0.5 for open/close, and the coordinate system is consistent with that of the CALVIN dataset.

\textbf{Space Alignment:} \textit{RoboData} includes the ML-45 version, which consists of 45 distinct tasks. To address the lack of training data for simulation learning, we adopt the scripted policies from Yu et al.~\cite{yu2020meta} and introduce Gaussian noise \(N(0, 0.1)\) to the generated actions at each step, resulting in a total of 22,500 trajectories, with each task producing 500 successful trajectories. To align with \textit{RoboData}'s predefined settings, we extract observations from the corner2 and gripperPOV perspectives. The rotational variables in the actions are zero-padded, and the gripper states are represented using -1 for open and 1 for closed, while other parameters remain unchanged.

For performance evaluation, we test on 20 unseen start and goal configurations for each task, totaling 900 unseen configurations. We report the average performance over these 900 trajectories, providing a comprehensive measure of the model's ability to generalize to new tasks and configurations.

\subsection{LIBERO Dataset}
LIBERO is a lifelong learning benchmark that includes multiple task suites involving various language-labeled rigid and articulated-body manipulation tasks. The dataset consists of a total of 130 tasks and 6,500 trajectories. The simulator includes four perspectives: frontview, birdview, agentview, sideview, all with a resolution of 256 × 256 pixels. The action representation differs from that used in CALVIN, employing CRMM to define actions.

\textbf{Space Alignment:} \textit{RoboData} includes the LIBERO-90 suite, which consists of 90 manipulation tasks, each with 50 demonstration trajectories collected through human teleoperation, providing a rich set of examples for training and evaluation. We select frontview and birdview as the observation perspectives, and camera parameters are obtained through replay. The coordinate system is defined as Forward-Left-Up. Due to differences in the coordinate system and workspace compared to the predefined settings in \textit{RoboData}, we align them through rotation and translation:
$$
W_{LIBERO} = \left[\begin{array}{cccc}
0 & 1 & 0 & 0.3 \\
-1 & 0 & 0 & 0 \\
0 & 0 & 1 & -0.1 \\
0 & 0 & 0 & 1 
\end{array}\right] W_{LIBERO}^{ori}.
$$
The states of the gripper are similarly represented using -1 for open and 1 for closed.

During evaluation, we test on 20 unseen start and goal configurations for each task, totaling 1,800 unseen configurations. This approach allows for a comprehensive assessment of the agent's performance and generalization capabilities, ensuring that the evaluation reflects the agent's ability to adapt to new situations and previously unencountered scenarios.

\subsection{RoboMimic  Dataset}
RoboMimic is a large-scale robotic manipulation benchmark designed to study imitation learning and offline reinforcement learning. The dataset includes 5 distinct manipulation tasks, each with a dataset of demonstrations teleoperated by proficient humans. These tasks are designed to enhance the learning effectiveness of robots through real human demonstrations.

\textbf{Space Alignment:} \textit{RoboData} includes 3 of these tasks (Lift, Can, Square) and excludes the other two dual-arm tasks. Given that the characteristics of RoboMimic align with those of LIBERO, all alignment methods can refer to LIBERO.

During evaluation, we test on 20 unseen start and goal configurations for each task, totaling 600 unseen configurations.

\subsection{RoboCasa  Dataset}
RoboCasa is an open-source simulation benchmark designed to study robotic manipulation tasks in household environments, utilizing a 12-DOF Franka robot, where the first 7 degrees of freedom are related to manipulation and the remaining 5 are related to mobility. The dataset includes a simulation environment featuring 120 distinct real-world scenes, thousands of interactive objects, and household appliances, utilizing generative AI tools to create environmental textures and 3D objects. The RoboCasa dataset introduces 100 systematic evaluation tasks, consisting of 25 atomic tasks and 75 composite tasks generated with the guidance of large language models. Additionally, RoboCasa provides a large-scale multi-task dataset containing over 100,000 trajectories for model training, showcasing performance improvements achieved through behavior cloning training with synthetic data, as well as the applicability of simulation data in real-world tasks. These features make RoboCasa an important resource for researching and developing language-conditioned robotic technologies, laying a solid foundation for advancing intelligent applications of robots in household environments.

\textbf{Space Alignment:} \textit{RoboData} includes 5,000 samples collected through remote control, utilizing two perspectives: front view and eye-in-hand. Only the degrees of freedom related to manipulation are retained. Given that the characteristics of RoboCasa align with those of LIBERO, all alignment methods can refer to LIBERO.

During evaluation, we test on 20 unseen start and goal configurations for each task, totaling 2,000 unseen configurations.

\subsection{ManiSkill2  Dataset}
ManiSkill2 is a unified benchmark designed for learning generalizable robotic manipulation skills, built on the SAPIEN platform. It includes 20 out-of-the-box task families, featuring over 2,000 distinct object models and more than 4 million demonstration frames. The dataset supports fast visual input learning algorithms, enabling a CNN-based policy to collect samples at approximately 2,000 frames per second (FPS) using just one GPU and 16 processes on a workstation. As the next generation of the SAPIEN ManiSkill benchmark, ManiSkill2 addresses critical pain points often encountered by researchers when utilizing benchmarks for developing generalizable manipulation skills, covering various task types, including stationary/mobile bases, single/dual-arm, and rigid/soft-body manipulation tasks. This extensive diversity of tasks and objects aims to enhance the robustness and applicability of robotic manipulation algorithms in real-world scenarios, making it an essential resource for advancing research in the field.

\textbf{Space Alignment:} \textit{RoboData} includes 20 tasks related to single-arm manipulation from the ManiSkill2 dataset. The coordinate system and workspace are defined as Forward-Right-Down and [-0.26, -0.79, -1.17] to [0.85, 0.76, 0.00]. To ensure spatial consistency and compatibility, the corresponding coordinate transformations are applied:
$$
W_{ManiSkill2} = \left[\begin{array}{cccc}
0 & 1 & 0 & 0.3 \\
1 & 0 & 0 & 0 \\
0 & 0 & -1 & 0 \\
0 & 0 & 0 & 1 
\end{array}\right] W_{ManiSkill2}^{ori}.
$$
The action representation uses CRMM to replace PCM. Since the original data did not include the camera's intrinsic and extrinsic parameters, we replayed the data and saved the relevant parameters.

During evaluation, we test on 20 unseen start and goal configurations for each task, totaling 400 unseen configurations.

\subsection{RoboCAS  Dataset}
RoboCAS is a benchmark proposed by Meituan's Embodied Intelligence Team, specifically designed for complex object arrangement scenarios in robotic manipulation. It is the first benchmark of its kind for such tasks and the first to employ a flexible and concise scripting strategy to collect samples in a cost-effective and efficient manner. RoboCAS showcases the handling of dispersed, ordered, and stacked objects within a highly realistic physical simulation environment, aiming to enhance robots' operational capabilities and performance across diverse settings. The benchmark provides a variety of proprioceptive observations and visual data, including RGB images and depth maps captured from the left gripper camera, base camera, and static camera.

\textbf{Space Alignment:} \textit{RoboData} includes all samples, utilizing only the base camera and static camera. The coordinate system and workspace are defined as Forward-Left-Up and [-0.70, -0.82, 0.062] to [0.85, 0.67, 0.92]. To ensure spatial consistency and compatibility, the following coordinate transformation is applied:
$$
W_{RLBench} = \left[\begin{array}{cccc}
0 & 1 & 0 & 0.3 \\
-1 & 0 & 0 & 0 \\
0 & 0 & 1 & 0.7 \\
0 & 0 & 0 & 1 
\end{array}\right] W_{RLBench}^{ori}.
$$
Since only end-effector positions are provided in the dataset, the research team utilized a composite rotation matrix to generate corresponding action representations, changing the gripper's open/close state from 0/0.04 to -1/1. Notably, the RGB images from this perspective are 480x640 pixels; to maintain consistency across all data in \textit{RoboData}, we only extract the central region of 480x480 pixels.

During evaluation, we test on 20 unseen start and goal configurations for each task.

\subsection{RLBench  Dataset}
RLBench is a challenging benchmark and learning environment specifically designed for robot learning. This benchmark features 18 completely unique, hand-designed tasks that range in difficulty from simple target reaching and door opening to more complex multi-stage tasks, such as opening an oven and placing a tray inside. RLBench provides a variety of proprioceptive observations and visual observation data, including RGB images, depth maps, and segmentation masks from the left shoulder, right shoulder, wrist, and front views.

\textbf{Space Alignment:} \textit{RoboData} includes all experiments, totaling 1.8 experiments, with visual input extracted from the wrist and front views. The coordinate system in this dataset differs from that of other datasets, defined as Forward-Left-up, with a workspace range from [-0.89, -0.72, 0.80] to [0.56, 0.69, 1.89]. We apply spatial transformations to shift the data into a predefined coordinate system. 
$$
W_{RLBench} = \left[\begin{array}{cccc}
0 & 1 & 0 & 0 \\
-1 & 0 & 0 & 0 \\
0 & 0 & 1 & 0.7 \\
0 & 0 & 0 & 1 
\end{array}\right] W_{RLBench}^{ori}.
$$
Additionally, only end-effector positions are provided, so we use CRMM to transform action representations, changing the gripper's open/close state from 0/1 to -1/1.

During evaluation, we test on 20 unseen start and goal configurations for each task, totaling 360 unseen configurations.

\subsection{Colosseum  Dataset}
Colosseum is a benchmark that complements RLBench by addressing the limitations of single environmental variables. It features 20 diverse manipulation tasks that enable systematic evaluation of models across 14 axes of environmental perturbations. These perturbations include changes in the color, texture, and size of objects, as well as variations in tabletop surfaces, backgrounds, and the physical properties of objects. Additionally, lighting conditions, distractors, and camera poses are adjusted. All configurations align with those of RLBench, allowing researchers to test and compare the robustness and adaptability of their models under a wider range of environmental conditions.

\textbf{Space Alignment:} \textit{RoboData} includes all samples, and the alignment method is consistent with that of RLBench.

\section{Comparison with OpenVLA}

To evaluate the superiority of our model architecture, we compare \iRoboMM{} with the currently best-performing OpenVLA. To ensure a fair comparison, we set the window size to 1, and train \iRoboMM{} from scratch, while OpenVLA is fine-tuned on the officially released weights.

As shown in Table~\ref{tab::Advanced}, \iRoboMM{} outperforms OpenVLA (LoRA) in multiple metrics, particularly in Task 3-5 and average sequence length. This indicates that \iRoboMM{} can capture longer dependencies when handling tasks, thereby improving model accuracy. These results not only demonstrate the superior performance of the \iRoboMM{} architecture but also provide valuable references for future research.

\begin{table}[ht]
    \small
    \centering
    \begin{adjustbox}{width=0.46\textwidth}
        \begin{tabular}{@{}c@{\hspace{5pt}}|c@{\hspace{5pt}} c@{\hspace{5pt}} c@{\hspace{5pt}} c@{\hspace{5pt}} c@{\hspace{1pt}} c@{} }
            \hline
            & \multicolumn{5}{c}{Task Completed in a Sequence } &   \multirow{2}{*}{Avg Len} \\
            \cline{2-6}  &  1  &   2 &   3 &   4 &   5 &  \\
            \hline
            OpenVLA (LoRA) & 78\% & 55\% & 29\% & 17\% & 8\% & 1.86 \\
            \hline
            \iRoboMM (ours) & 81\% & 54\% & 37\% & 25\% & 16\% & 2.15 \\
            \hline
        \end{tabular}
    \end{adjustbox}
    \caption{Performance comparison with OpenVLA on CALVIN.}
    \label{tab::Advanced}
    \vspace{-4mm}
\end{table}

\section{Experiment Details}
The success rates of the expert models in the Table~\ref{tab:sota} are organized from the following sources:
The evaluation methods on the CALVIN~\cite{mees2022calvin} dataset are sourced from the official CALVIN leaderboard (url: http://calvin.cs.uni-freiburg.de/).
In the Meta-World~\cite{yu2020meta} dataset, the results of PAD~\cite{guo2024prediction}, GR-1~\cite{wu2023unleashing}, SuSIE~\cite{black2023zero}, RT-2*~\cite{brohan2023rt}, and RT-1~\cite{brohan2022rt} come from PAD~\cite{guo2024prediction}, while the results of PRISE~\cite{zheng2024prise} are derived from related papers.
In the Libero~\cite{liu2024libero} dataset, the results of QueST~\cite{mete2024questselfsupervisedskillabstractions}, VQ-BeT~\cite{lee2024behavior}, PRISE~\cite{zheng2024prise},  DiffusionPolicy~\cite{chi2023diffusion}, ACT~\cite{zhao2023learning}, and ResNet-T~\cite{liu2024libero} all come from QueST~\cite{mete2024questselfsupervisedskillabstractions}, while the results of MDT~\cite{reuss2024multimodal} and Distill-D~\cite{ha2023scaling} are sourced from MDT~\cite{reuss2024multimodal}; the results of MaIL~\cite{jia2024mail}, ATM~\cite{wen2023atm}, and MUTEX~\cite{shah2023mutex} come from their respective research papers.
The results of RoboCasa~\cite{robocasa2024} in the RoboCasa~\cite{robocasa2024} dataset are sourced from related papers.
In the Rt-1~\cite{brohan2022rt} dataset, all results come from CogACT~\cite{li2024cogact} paper.

The success rates of each task for \textit{RoboData} on various datasets are shown in the Table~\ref{tab:ur_calvin}, ~\ref{tab:ur_metaworld}, ~\ref{tab:rt-1}, ~\ref{tab:ur_libero}, ~\ref{tab:ur_robocasa}.

\begin{table}[ht]
    \small
    \centering
    \begin{tabular}{@{}lc@{}} 
        \toprule
        \textbf{Task Name} & \textbf{Success Rate} \\ \midrule
        CoffeePressButton & 100\% \\
        CoffeeServeMug & 50\% \\
        CoffeeSetupMug & 25\% \\
        CloseDoubleDoor & 30\% \\
        CloseSingleDoor & 100\% \\
        OpenDoubleDoor & 0\% \\
        OpenSingleDoor & 45\% \\
        CloseDrawer & 100\% \\
        OpenDrawer & 80\% \\
        TurnOffMicrowave & 75\% \\
        TurnOnMicrowave & 85\% \\
        PnPCabToCounter & 45\% \\
        PnPCounterToCab & 50\% \\
        PnPCounterToMicrowave & 35\% \\
        PnPCounterToSink & 40\% \\
        PnPCounterToStove & 60\% \\
        PnPMicrowaveToCounter & 70\% \\
        PnPSinkToCounter & 40\% \\
        PnPStoveToCounter & 40\% \\
        TurnOffSinkFaucet & 70\% \\
        TurnOnSinkFaucet & 55\% \\
        TurnSinkSpout & 90\% \\
        TurnOffStove & 35\% \\
        TurnOnStove & 55\% \\
        PrepareCoffee & 0\% \\
        ArrangeVegetables & 0\% \\
        MicrowaveThawing & 0\% \\
        RestockPantry & 0\% \\
        PreSoakPan & 0\% \\
        \bottomrule
    \end{tabular}
    \caption{\iRoboMM{} Success Rates on Various Tasks in RoboCasa~\cite{robocasa2024}.}
    \label{tab:ur_robocasa}
\end{table}

\begin{table}[!ht]
    \small
    \centering
    \begin{tabular}{@{}ll@{}}
        \toprule
        \textbf{Task} & \textbf{Success Rate} \\ 
        \midrule
        rotate\_blue\_block\_right & 82.6\% \\
        move\_slider\_right & 98.2\% \\
        lift\_red\_block\_slider & 87.9\% \\
        place\_in\_slider & 26.9\% \\
        turn\_off\_lightbulb & 89.7\% \\
        turn\_off\_led & 96.9\% \\
        push\_into\_drawer & 68.1\% \\
        lift\_blue\_block\_drawer & 100.0\% \\
        lift\_pink\_block\_slider & 86.2\% \\
        open\_drawer & 93.1\% \\
        lift\_pink\_block\_table & 87.0\% \\
        turn\_on\_lightbulb & 94.4\% \\
        rotate\_blue\_block\_left & 96.6\% \\
        push\_blue\_block\_left & 87.1\% \\
        close\_drawer & 91.0\% \\
        rotate\_red\_block\_right & 80.3\% \\
        push\_pink\_block\_right & 64.9\% \\
        push\_red\_block\_right & 59.7\% \\
        push\_red\_block\_left & 87.1\% \\
        lift\_blue\_block\_table & 87.2\% \\
        place\_in\_drawer & 97.3\% \\
        move\_slider\_left & 91.9\% \\
        rotate\_red\_block\_left & 84.6\% \\
        turn\_on\_led & 93.0\% \\
        lift\_red\_block\_table & 93.0\% \\
        stack\_block & 55.4\% \\
        push\_pink\_block\_left & 91.2\% \\
        lift\_blue\_block\_slider & 85.2\% \\
        unstack\_block & 100.0\% \\
        rotate\_pink\_block\_left & 90.2\% \\
        lift\_pink\_block\_drawer & 85.7\% \\
        rotate\_pink\_block\_right & 63.5\% \\
        lift\_red\_block\_drawer & 93.3\% \\
        push\_blue\_block\_right & 48.4\% \\
        \bottomrule
    \end{tabular}
    \caption{\iRoboMM{} Success Rates on Various Tasks in CALVIN~\cite{mees2022calvin}.}
    \label{tab:ur_calvin}
\end{table}

\begin{table}[ht]
    \small
    \centering
    \begin{tabular}{@{}ll@{}}
        \toprule
        \textbf{Task} & \textbf{Success Rate} \\ 
        \midrule
        assembly-v2 & 100\% \\
        basketball-v2 & 100\% \\
        bin-picking-v2 & 70\% \\
        box-close-v2 & 85\% \\
        button-press-topdown-v2 & 100\% \\
        button-press-topdown-wall-v2 & 100\% \\
        button-press-v2 & 80\% \\
        button-press-wall-v2 & 85\% \\
        coffee-button-v2 & 90\% \\
        coffee-pull-v2 & 40\% \\
        coffee-push-v2 & 65\% \\
        dial-turn-v2 & 100\% \\
        disassemble-v2 & 80\% \\
        door-close-v2 & 100\% \\
        door-lock-v2 & 100\% \\
        door-open-v2 & 100\% \\
        door-unlock-v2 & 100\% \\
        hand-insert-v2 & 55\% \\
        drawer-close-v2 & 100\% \\
        drawer-open-v2 & 100\% \\
        faucet-open-v2 & 0\% \\
        faucet-close-v2 & 90\% \\
        hammer-v2 & 15\% \\
        handle-press-side-v2 & 100\% \\
        handle-press-v2 & 100\% \\
        handle-pull-side-v2 & 25\% \\
        handle-pull-v2 & 100\% \\
        lever-pull-v2 & 80\% \\
        peg-insert-side-v2 & 55\% \\
        pick-place-wall-v2 & 95\% \\
        pick-out-of-hole-v2 & 15\% \\
        reach-v2 & 75\% \\
        push-back-v2 & 100\% \\
        push-v2 & 90\% \\
        pick-place-v2 & 100\% \\
        plate-slide-v2 & 100\% \\
        plate-slide-side-v2 & 100\% \\
        plate-slide-back-v2 & 100\% \\
        plate-slide-back-side-v2 & 100\% \\
        peg-unplug-side-v2 & 25\% \\
        soccer-v2 & 20\% \\
        stick-push-v2 & 100\% \\
        stick-pull-v2 & 85\% \\
        push-wall-v2 & 100\% \\
        reach-wall-v2 & 85\% \\
        shelf-place-v2 & 45\% \\
        sweep-into-v2 & 95\% \\
        sweep-v2 & 100\% \\
        window-open-v2 & 60\% \\
        window-close-v2 & 100\% \\
        \bottomrule
    \end{tabular}
    \caption{\iRoboMM{} Success Rates on Various Tasks in Meta-World~\cite{yu2020meta}}
    \label{tab:ur_metaworld}
\end{table}

\begin{table}[ht]
    \vspace{-2.5mm}
    \centering
    \scriptsize
    \begin{tabular}{@{}l@{\hspace{8pt}}c@{\hspace{10pt}}c@{\hspace{10pt}}c@{\hspace{10pt}}c@{}}
        \toprule
        Method & Pick Coke Can & Move Near  & Open / Close Drawer   & Overall \\
        \midrule
        RT-1-X      & 0.567 & 0.317  & 0.597 & 0.534 \\
        \textcolor{gray}{RT-2-X(55B)}      & \textcolor{gray}{0.787} & \textcolor{gray}{0.779} & \textcolor{gray}{0.250} & \textcolor{gray}{0.607} \\
        Octo-Base   & 0.170 & 0.042 & 0.227 & 0.169 \\
        OpenVLA     & 0.163 & 0.462 & 0.356 & 0.248 \\
        HPT[1]      & 0.60  & 0.24  & 0.23  & 0.35 \\
        \midrule
        \iRoboMM      & 0.63  & 0.64  & 0.525 & 0.60\\
        \bottomrule
    \end{tabular}
    \vspace{-3.0mm}
    \caption{SIMPLER evaluation results of different methods on RT-1. The “Overall” column reports the success rate averaged across the sub-tasks of all task types.}
    \label{tab:rt-1}
    \vspace{-2.5mm}
\end{table}

\begin{table}[ht]
    \small
    \centering
    \begin{tabular}{@{}c@{\hspace{8pt}}c@{\hspace{8pt}}|c@{\hspace{8pt}}c@{}}
        \toprule
        \textbf{Task Index} & \textbf{Success Rate} & \textbf{Task Index} & \textbf{Success Rate} \\ \midrule
        0 & 100\% & 45 & 80\% \\
        1 & 85\% & 46 & 90\% \\
        2 & 95\% & 47 & 100\% \\
        3 & 95\% & 48 & 95\% \\
        4 & 85\% & 49 & 95\% \\
        5 & 60\% & 50 & 95\% \\
        6 & 100\% & 51 & 35\% \\
        7 & 100\% & 52 & 95\% \\
        8 & 90\% & 53 & 100\% \\
        9 & 80\% & 54 & 95\% \\
        10 & 100\% & 55 & 100\% \\
        11 & 95\% & 56 & 100\% \\
        12 & 90\% & 57 & 100\% \\
        13 & 75\% & 58 & 100\% \\
        14 & 95\% & 59 & 90\% \\
        15 & 95\% & 60 & 100\% \\
        16 & 100\% & 61 & 95\% \\
        17 & 95\% & 62 & 95\% \\
        18 & 95\% & 63 & 100\% \\
        19 & 100\% & 64 & 80\% \\
        20 & 100\% & 65 & 100\% \\
        21 & 85\% & 66 & 100\% \\
        22 & 90\% & 67 & 100\% \\
        23 & 70\% & 68 & 95\% \\
        24 & 100\% & 69 & 95\% \\
        25 & 100\% & 70 & 100\% \\
        26 & 90\% & 71 & 100\% \\
        27 & 75\% & 72 & 95\% \\
        28 & 100\% & 73 & 70\% \\
        29 & 100\% & 74 & 90\% \\
        30 & 100\% & 75 & 70\% \\
        31 & 100\% & 76 & 100\% \\
        32 & 50\% & 77 & 100\% \\
        33 & 85\% & 78 & 85\% \\
        34 & 100\% & 79 & 100\% \\
        35 & 85\% & 80 & 90\% \\
        36 & 90\% & 81 & 60\% \\
        37 & 100\% & 82 & 100\% \\
        38 & 80\% & 83 & 95\% \\
        39 & 85\% & 84 & 90\% \\
        40 & 95\% & 85 & 95\% \\
        41 & 100\% & 86 & 90\% \\
        42 & 100\% & 87 & 100\% \\
        43 & 100\% & 88 & 100\% \\
        44 & 95\% & 89 & 95\% \\ 
        \bottomrule
    \end{tabular}
    \caption{\iRoboMM{} Success Rates on Various Tasks in Libero~\cite{liu2024libero}.}
    \label{tab:ur_libero}
\end{table}


\end{document}